\documentclass{article}
\usepackage[final]{neurips_2025}

\usepackage[utf8]{inputenc} 
\usepackage[T1]{fontenc}    
\usepackage{hyperref}       
\usepackage{url}            
\usepackage{booktabs}       
\usepackage{amsfonts}       
\usepackage{nicefrac}       
\usepackage{microtype}      
\usepackage{xcolor}         
\usepackage{graphicx,subfigure} 
\usepackage{natbib}  
\setcitestyle{numbers,square}

\usepackage{amsmath}
\usepackage[linesnumbered,algoruled,boxed,lined]{algorithm2e}
\usepackage{colortbl,multirow,rotating,makecell}
\usepackage{wrapfig}
\usepackage[most]{tcolorbox}

\newcommand{\methodname}{{\tt{SV-NUP}}}

\title{Efficient Shapley Value-based Non-Uniform Pruning of Large Language Models}

\author{%
  Chuan Sun, Han Yu\\
  College of Computing and Data Science\\
  Nanyang Technological University\\
  Singapore\\
  \texttt{\{chuan.sun, han.yu\}@ntu.edu.sg} \\
  \And
  Lizhen Cui \\
  School of Software\\
  Shandong University \\
  China \\
  \texttt{clz@sdu.edu.cn} \\
  \AND
  Xiaoxiao Li \\
  Department of Electrical and Computer Engineering\\
  The University of British Columbia \\
  Canada \\
  \texttt{xiaoxiao.li@ece.ubc.ca} \\
}

\begin{document}

\maketitle

\begin{abstract}
Pruning large language models (LLMs) is a promising solution for reducing model sizes and computational complexity while preserving performance. Traditional layer-wise pruning methods often adopt uniform sparsity at all Transformer layers, which leads to suboptimal performance due to the varying significance of transformer layers not being accounted for. To this end, we propose the \underline{S}hapley \underline{V}alue-based \underline{N}on-\underline{U}niform \underline{P}runing (\methodname{}) method for LLMs. This method quantifies the contribution of each transformer layer to the overall model performance, enabling the assignment of tailored pruning budgets to different layers to retain critical parameters. To further improve efficiency, we design the \underline{S}liding \underline{W}indow-based \underline{S}hapley \underline{V}alue (SWSV) approximation method. It substantially reduces computational overhead compared to exact SV calculation methods. Extensive experiments on various LLMs including LLaMA-v1/v2/v3, and OPT demonstrate the effectiveness of \methodname{}. The results reveal that non-uniform pruning significantly enhances the performance of pruned models. Notably, \methodname{} achieves a reduction in perplexity (PPL) of 18.01\% and 19.55\% on LLaMA-7B and LLaMA-13B, respectively, compared to SparseGPT at 70\% sparsity.
\end{abstract}

\section{Introduction}
Large language models (LLMs) have emerged as a transformative technology, demonstrating remarkable capabilities in natural language understanding and generation tasks \cite{brown2020language}. LLMs have showcased significant adaptability through fine-tuning, enabling their deployment in highly specialized applications. These advantages underscore the critical role of LLMs in solving real-world challenges \cite{patterson2021carbon}. Despite these capabilities, the deployment of LLMs has been hindered by their immense computational demands. Modern LLMs often consist of billions or even trillions of parameters \cite{luccioni2023estimating}, as seen in models like GPT-3 (175 billion parameters) and PaLM (540 billion parameters). The immense scale incurs significant memory, storage and power costs, making it challenging to run these models on resource-constrained devices. To address these issues, researchers have increasingly turned to model compression techniques, particularly pruning, to reduce the size and computational requirements of LLMs, while retaining their performance \cite{frantar2023sparsegpt}.

Pruning \cite{lecun1989optimal,han2015learning,han2015deep} is one of the most prominent model compression techniques. It aims to remove redundant or less important parameters from neural networks, thereby reducing their sizes and computational complexity. Recent advances in pruning have progressed from unstructured sparsity (i.e., individual weight removal) to structured sparsity (i.e., eliminating entire neurons, heads or layers), thereby enabling hardware-efficient implementation \cite{frantar2023sparsegpt,jaiswal2024emergence,ma2023llm,xia2023sheared}. For LLMs, several state-of-the-art methods have been proposed, including magnitude pruning, lottery ticket hypothesis approaches, and structured pruning based on attention mechanisms. These methods demonstrate the potential of pruning to enable LLM deployment on resource-constrained devices without causing substantial performance degradation.

\begin{figure}[t]
\centering
\includegraphics[width=1\linewidth]{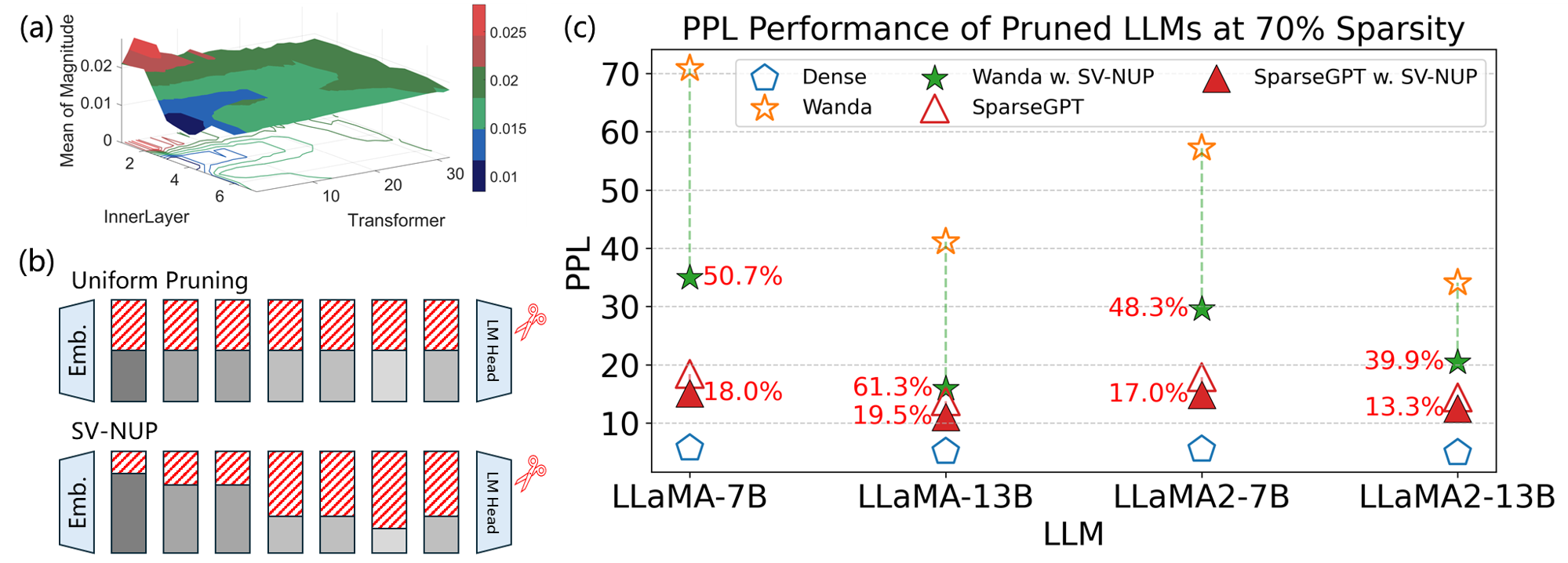}
\caption{Overview and effectiveness of \methodname{}: (a) Distribution of weight magnitude across Transformer layers; (b) Conceptual comparison of uniform pruning vs. \methodname{};
(c) PPL\textdownarrow~ (the smaller the better) comparison under different pruning strategies. The red number is the improvement percentage. Unlike uniform pruning, which ignores layer-wise importance and may degrade LLM performance, \methodname{} leverages Shapley value to estimate the contribution of each Transformer layer and allocates pruning ratios accordingly to better preserve model performance.}
\label{fig:motivation}
\vspace{-0.1in}
\end{figure}

Existing pruning methods often use a layer-wise strategy that applies uniform sparsity across all layers, ignoring their varying importance. While simple to implement, this approach overlooks the inherent differences in the contributions of different layers to the overall performance of the model. Thus, it can only find the local optimal pruning solution, but not the global optimal solution. Empirical evidence also suggests that certain layers are more critical than others, and uniformly pruning across all layers may lead to the removal of essential parameters, ultimately impairing the pruned model's performance \cite{mocanu2018scalable,yin2023outlier}.

To address this limitation, we leverage the concept of Shapley value from cooperative game theory and propose a novel \underline{S}hapley \underline{V}alue-based \underline{N}on-\underline{U}niform \underline{P}runing (\methodname{}) method for LLMs,  which was originally designed to fairly assess the contributions of multiple players in a game. In this study, we treat each Transformer layer in an LLM as a ``player". Based on this idea, we evaluate the contribution of each Transformer layer to the overall performance of an LLM. The results are further used as a basis to allocate customized sparsity ratios to each Transformer layer, prioritizing the preservation of parameters in more important ones. This approach not only improves the performance of the pruned LLMs, but also introduces a theoretical foundation for sparsity allocation, moving beyond heuristic methods. We further propose a \underline{S}liding \underline{W}indow-based \underline{S}hapley \underline{V}alue approximation method (SWSV) to reduce the computational cost. The overview of \methodname{} is shown in Figure \ref{fig:motivation}.

We conduct extensive experiments to evaluate the performance of \methodname{} for pruning LLMs across a spectrum of LLMs, including LLaMA-v1/v2/v3, and OPT. \methodname{} achieves a reduction in perplexity (PPL) of 18.01\% and 19.55\% on LLaMA-7B and LLaMA-13B, respectively, compared to SparseGPT at 70\% sparsity. The results validate our hypothesis that non-uniform sparsity allocation can significantly enhance the performance of pruned models, paving the way for more efficient and practical LLM deployment on resource-constrained devices.

\section{Related Work}
LLM pruning often requires retraining to recover performance, which presents substantial challenges. While existing LLM-specific pruning methods primarily adopt uniform pruning strategies \cite{ma2023llm,hu2021lora,frantar2023sparsegpt,sun2023simple}, non-uniform layerwise sparsity has been extensively studied in vision models \cite{mocanu2016topological,erdds1959random, mocanu2018scalable,evci2020rigging,liu2022unreasonable,frankle2018lottery,lee2018snip,wang2020picking,liu2021sparse}. However, global pruning is often inefficient and computationally expensive for LLMs. Although prior work has analyzed the importance of LLM components \cite{shim2022understanding,gromov2024unreasonable,michel2019sixteen,clark2019does,tenney2019bert,tenney2019you,zhang2024investigating}, none have explored Shapley value (SV)-based non-uniform pruning. \methodname{} fills this gap. For a more detailed literature review, refer to Appendix \ref{sec:related}.

\section{Motivation}
\label{sec:motivation}

\begin{figure}[t]
\centering

\subfigure[TinyLLaMA with Mean] {\label{fig:mean-a}
\includegraphics[width=0.3\linewidth]{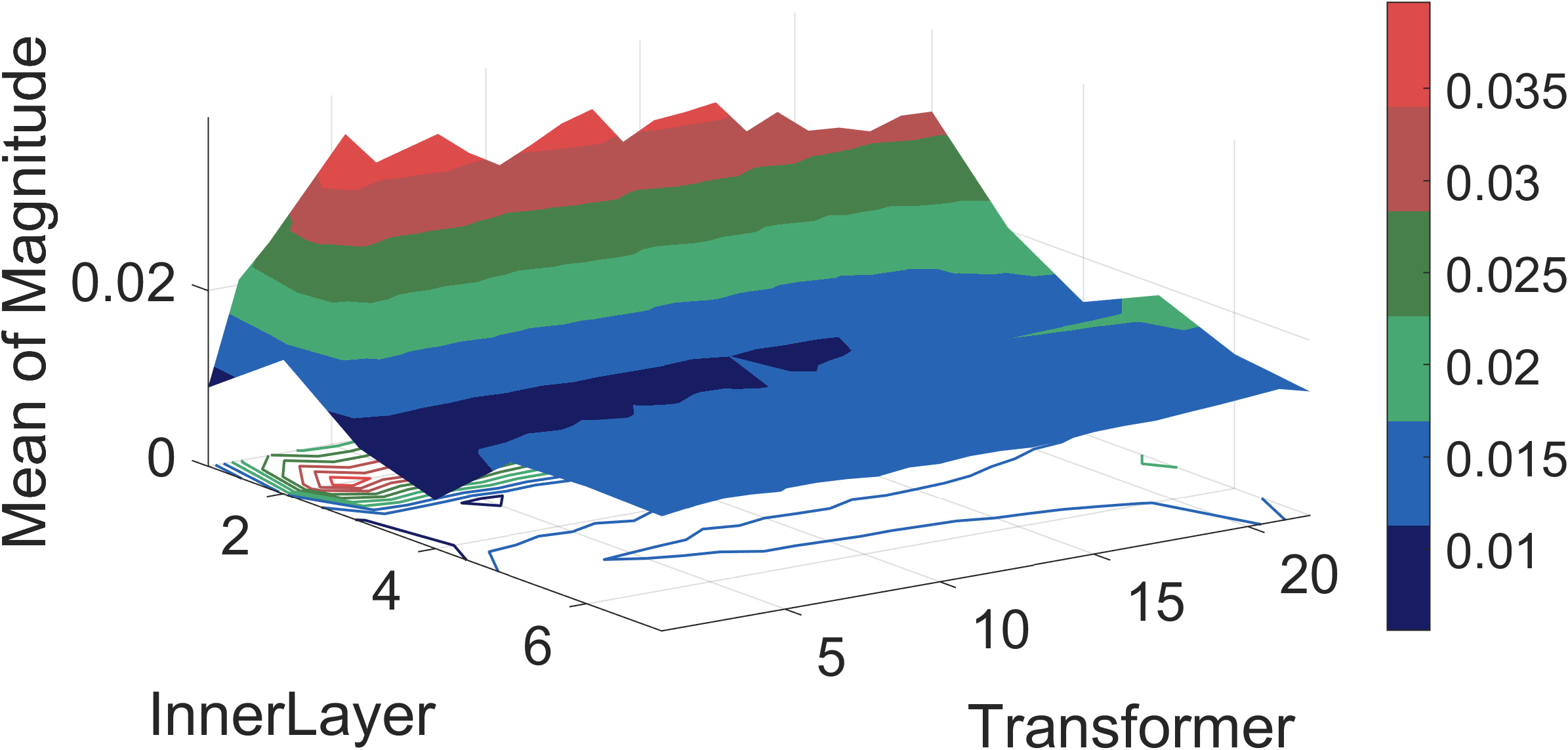}
}
\subfigure[LLaMA-7B with Mean] {\label{fig:mean-c}
\includegraphics[width=0.3\linewidth]{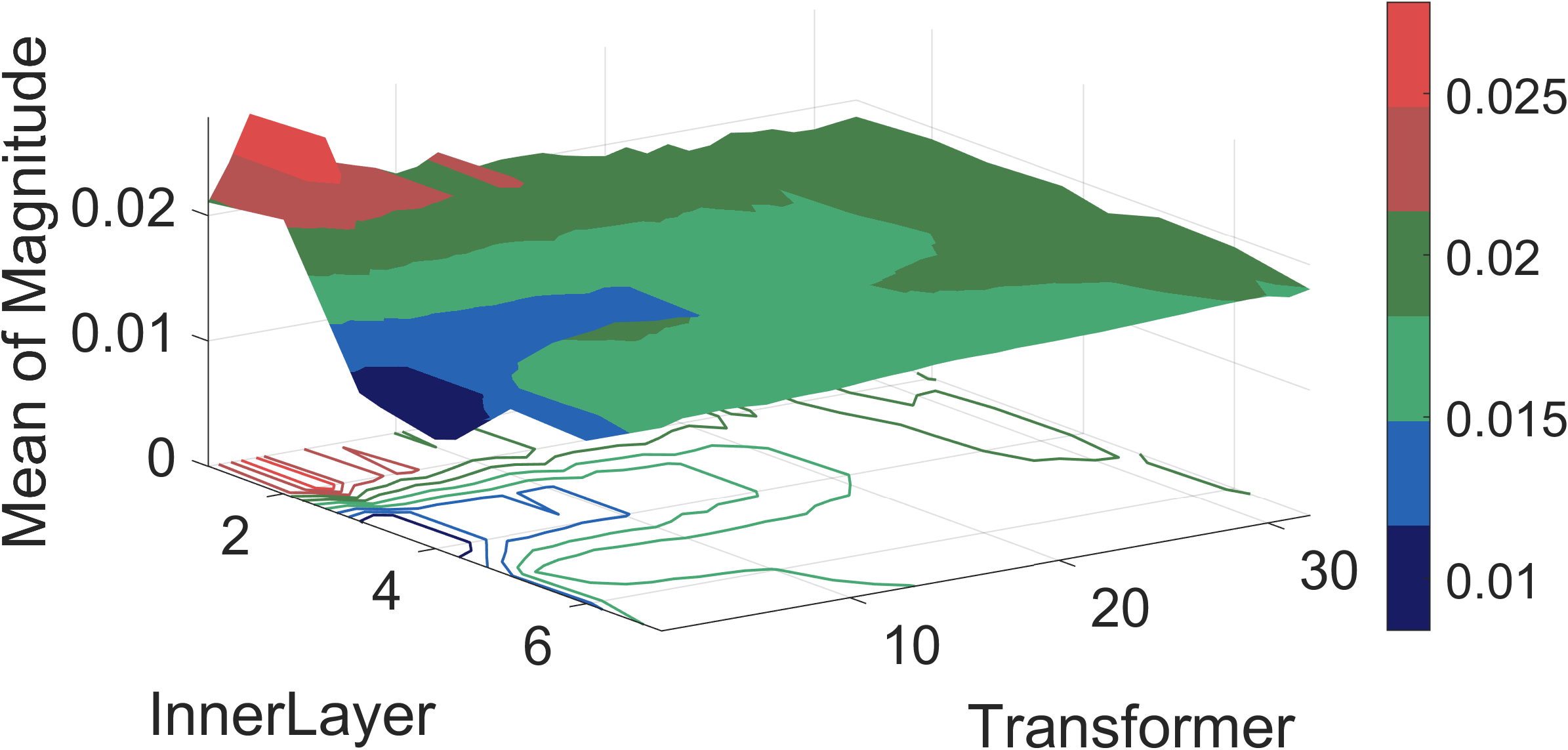}
}
\subfigure[Mistral-7B with Mean] {\label{fig:mean-g}
\includegraphics[width=0.3\linewidth]{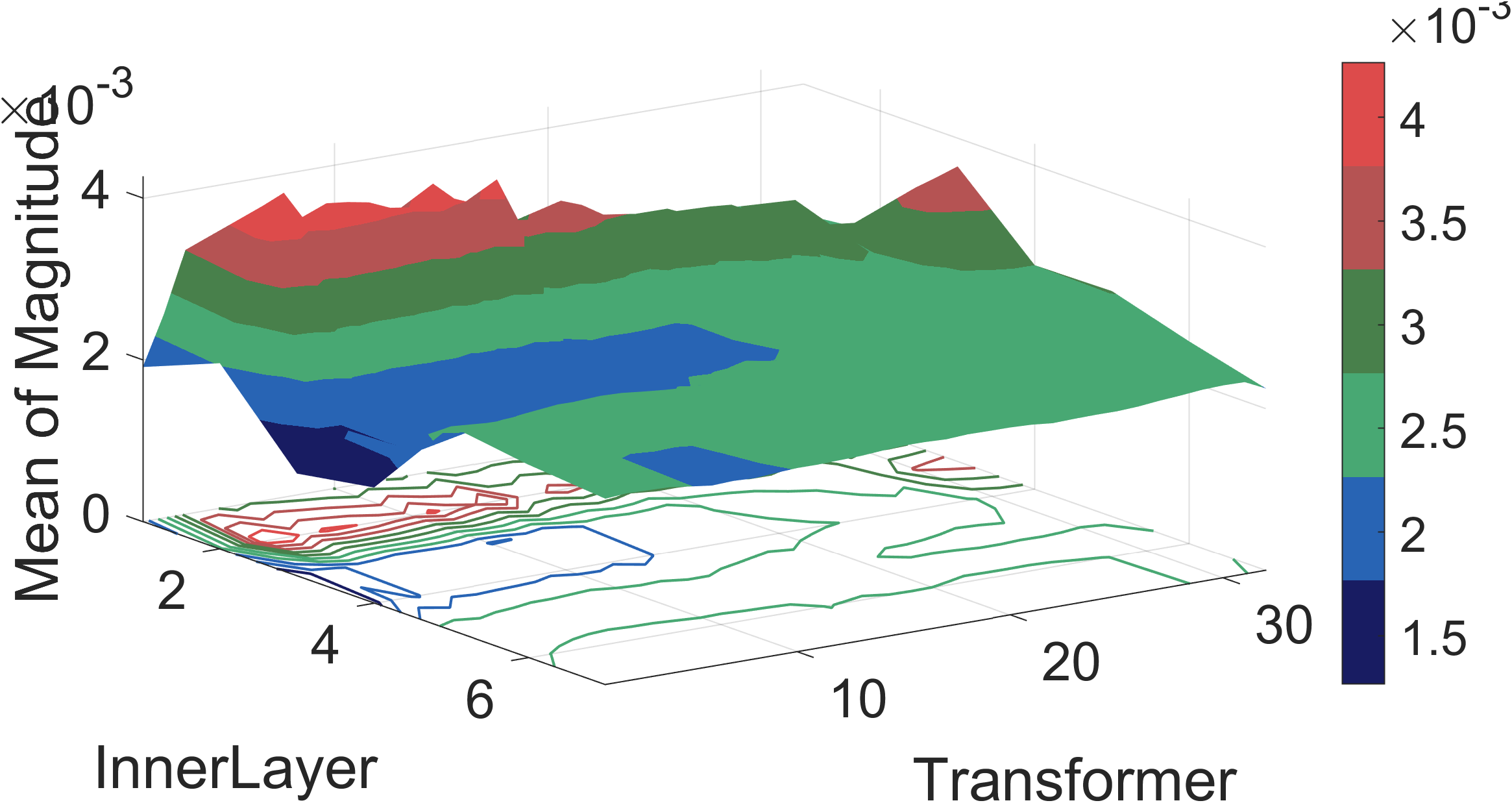}
}
\subfigure[TinyLLaMA with Std] {\label{fig:mean-b}
\includegraphics[width=0.3\linewidth]{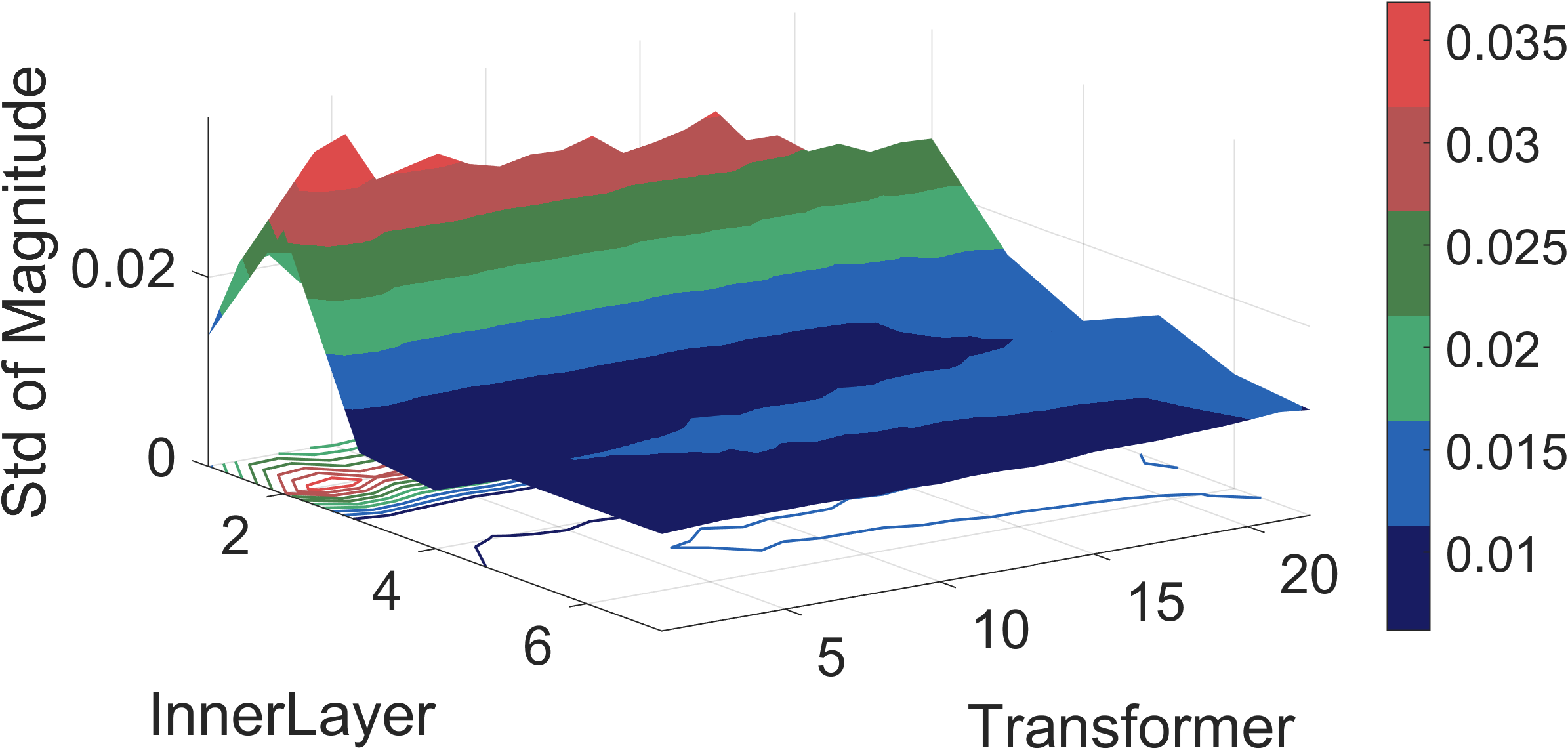}
}
\subfigure[LLaMA-7B with Std] {\label{fig:mean-d}
\includegraphics[width=0.3\linewidth]{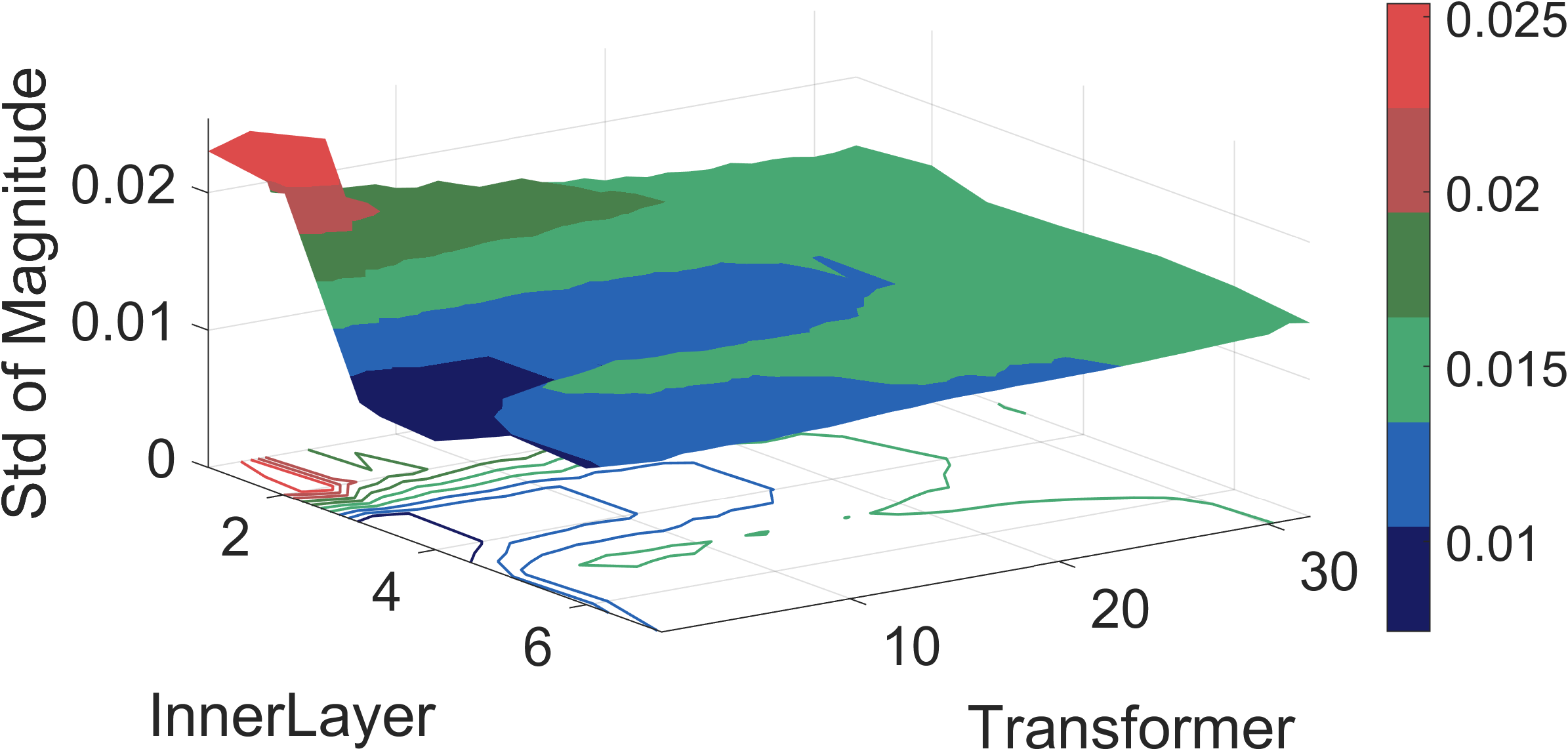}
}
\subfigure[Mistral-7B with Std] {\label{fig:mean-h}
\includegraphics[width=0.3\linewidth]{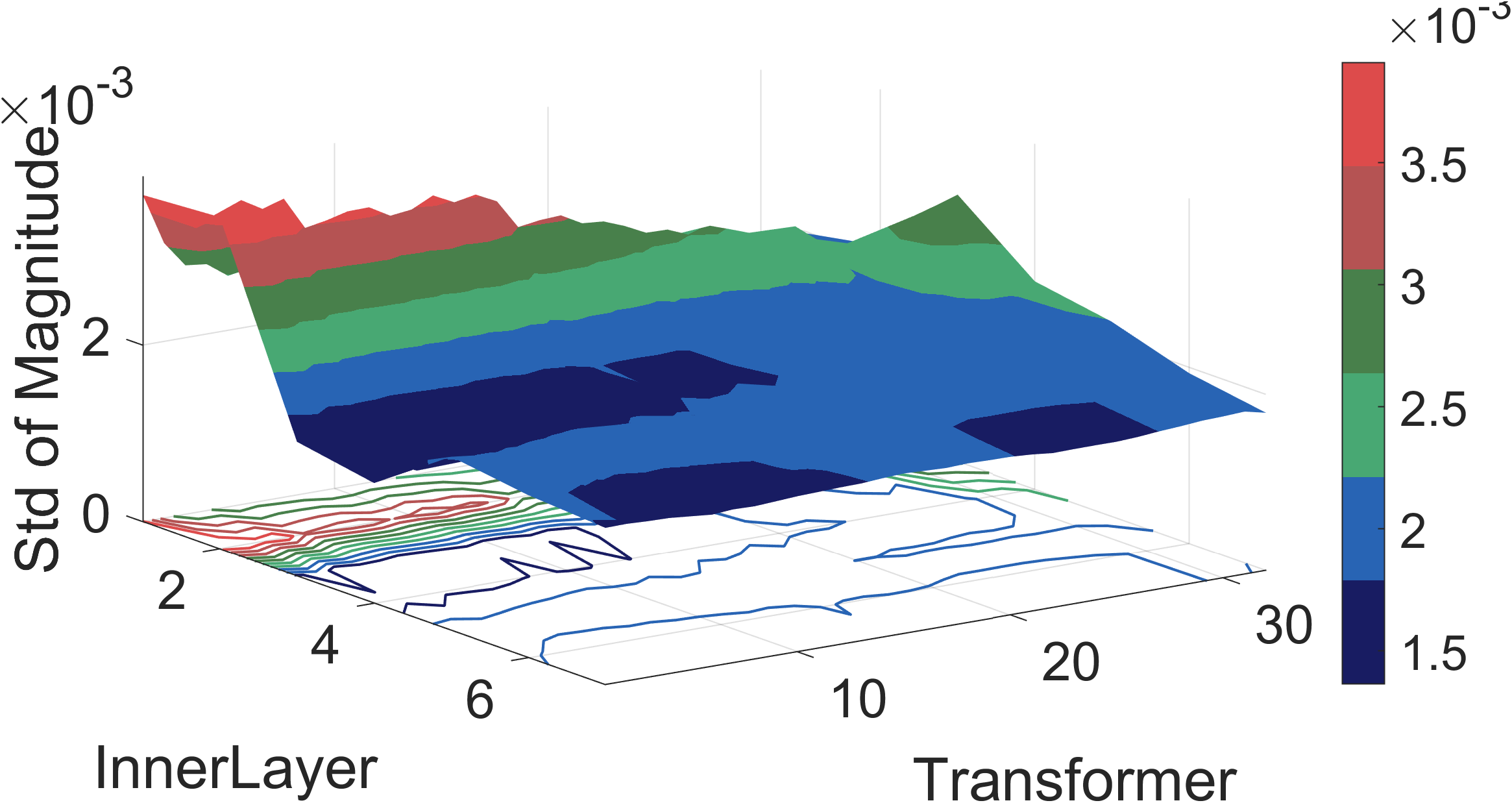}
}
\caption{Comparison of the mean and standard deviation (Std) across different Transformer layers under Magnitude. TinyLLaMA consists of 22 Transformer layers, LLama-7B and Mistral-7B consist of the same 32 Transformer layers, where each Transformer layer consists of 7 inner-layers.}
\label{fig:mean}
\vspace{-0.1in}
\end{figure}

In this section, we conduct empirical studies on three LLMs (TinyLLama, LLaMA-7B, and Mistral-7B) to motivate the need for allocating different pruning ratios across Transformer layers. We analyze the differences using the Magnitude method \cite{lee2020layer} and summarize each Transformer layer’s structure with the mean and standard deviation of innerlayer metrics. As shown in Figure \ref{fig:mean}, two key observations can be drawn from the results.

\subsection{Observations}

\begin{wrapfigure}{l}{0.43\textwidth}
\vspace{-0.2in}
\begin{minipage}{0.43\textwidth}
\begin{tcolorbox}[colback=black!5!white, colframe=black!70!white, title=Obervation 1]
Different Transformer layers in LLMs exhibit significant variations in the mean and std of magnitude.
\end{tcolorbox}
\end{minipage}
\vspace{-0.14in}
\end{wrapfigure}


As the Transformer index increases, the mean magnitude decreases across all LLMs (Figures \ref{fig:mean-a}, \ref{fig:mean-c} and \ref{fig:mean-g}). Notably, the changes in LLaMA-7B and Mistral-7B are more pronounced compared to TinyLLaMa. The earlier (shallower) Transformer layers tend to have higher mean magnitudes. From the std of magnitude in Figures \ref{fig:mean-b}, \ref{fig:mean-d} and \ref{fig:mean-h}, it can be observed that weights of shallow Transformer layers change dramatically. It is evident that the weights of shallow Transformer layers vary more significantly. Higher mean and std of magnitude values indicate that the corresponding Transformer layers have a greater impact on LLM performance. These results underscore that different layers contribute unevenly to the overall model performance, suggesting that some Transformer layers are more sensitive to changes in parameter magnitudes.

\begin{wrapfigure}{l}{0.43\textwidth}
\vspace{-0.2in}
\begin{minipage}{0.43\textwidth}
\begin{tcolorbox}[colback=black!5!white, colframe=black!70!white, title=Obervation 2]
The attention module of Transformer layers has a higher mean and std of magnitude than the feed-forward network (FFN) module.
\end{tcolorbox}
\end{minipage}
\vspace{-0.14in}
\end{wrapfigure}


Each Transformer layer consists of 7 inner layers. The first 4 correspond to the attention module, while the last 3 form the FFN module. In all the figures, the mean and std of magnitude in the attention module are higher than those in the FFN module. This finding indicates that the attention module is more sensitive compared to the FFN module to changes in parameter magnitude. For instance, in Figures \ref{fig:mean-a} and \ref{fig:mean-b}, TinyLLaMa's inner layer 4 shows a distinct dividing line: layers 1–4 form peaks, whereas layers 5–7 form troughs. The other three LLMs follow similar trends. As a result, the attention module warrants more attention during pruning.

\subsection{Analysis and Implications}
Conventional LLM pruning approaches apply uniform pruning across all Transformer layers, ignoring differences across Transformer layers. As shown in Figure \ref{fig:mean-c}, early Transformer layers exhibit lower mean magnitudes, suggesting that they can withstand more aggressive pruning; whereas the latter Transformer layers show higher mean magnitudes and std, indicating that they should be pruned more conservatively to maintain performance. Applying the same pruning ratio to Transformer layers with high mean magnitudes or high std as those with lower values might remove critical capabilities from important Transformer layers, while preserving too many parameters in less significant ones.

This analysis clearly demonstrates the advantage of allocating different pruning ratios to different Transformer layers. By tailoring the pruning ratios based on the unique characteristics of each Transformer layer, pruning strategies can be optimized to retain crucial information in the model. Such a non-uniform pruning approach enables more efficient compression where less critical Transformer layers undergo more aggressive pruning, while vital Transformer layers are preserved, thereby enhancing the performance of the pruned model.

\section{The Proposed \methodname{} Method}
\label{sec:method}

Motivated by these observations, we investigate the issue of adaptive sparsity for LLM pruning. The Shapley value, a concept derived from cooperative game theory, is widely used to fairly allocate contributions among multiple players in a game. This paper leverages the Shapley value to allocate pruning ratios across different Transformer layers. By treating each Transformer layer as a ``player" in LLMs, we can compute the contribution of each Transformer layer to the overall performance of an LLM. This approach will allow us to assign pruning ratios dynamically, ensuring that more important Transformer layers (i.e., those with higher Shapley values) are pruned less aggressively, while less critical Transformer layers receive higher pruning ratios. We propose \methodname{} to address this problem, which provides a theoretically grounded and equitable solution to non-uniform pruning. This method optimizes the trade-off between model size and performance, ensuring that critical Transformer layers are preserved, while less significant ones are pruned more aggressively.

We consider post-training pruning of LLMs from a well-optimized model with weights $\mathbf{W}^*$ to a sparse version $\mathbf{W}$ with many 0 under a given pruning ratio $\rho$. As LLMs contain hundreds of millions of parameters, we usually adopt layer-wise pruning. The neural network structure of LLMs is generally composed of many Transformer layers. We denote Transformer layers as the set $\mathcal{T}=\{1,2,...,T\}$. Conventional LLM pruning methods apply a uniform pruning ratio $\rho$ to all Transformer layers. According to the analysis in Section \ref{sec:motivation}, we should allocate different pruning ratios $\rho_t$ to different Transformer layer $t\in\mathcal{T}$. The key to allocating pruning ratios is to calculate the contribution of Transformer layers of LLMs precisely. We use SV to analyze each Transformer layer contribution.

\subsection{Transformer Layer Contribution by Shapley Value}

\begin{wrapfigure}{R}{0.6\textwidth}
\vspace{-0.1in}
\centering
\includegraphics[width=1\linewidth]{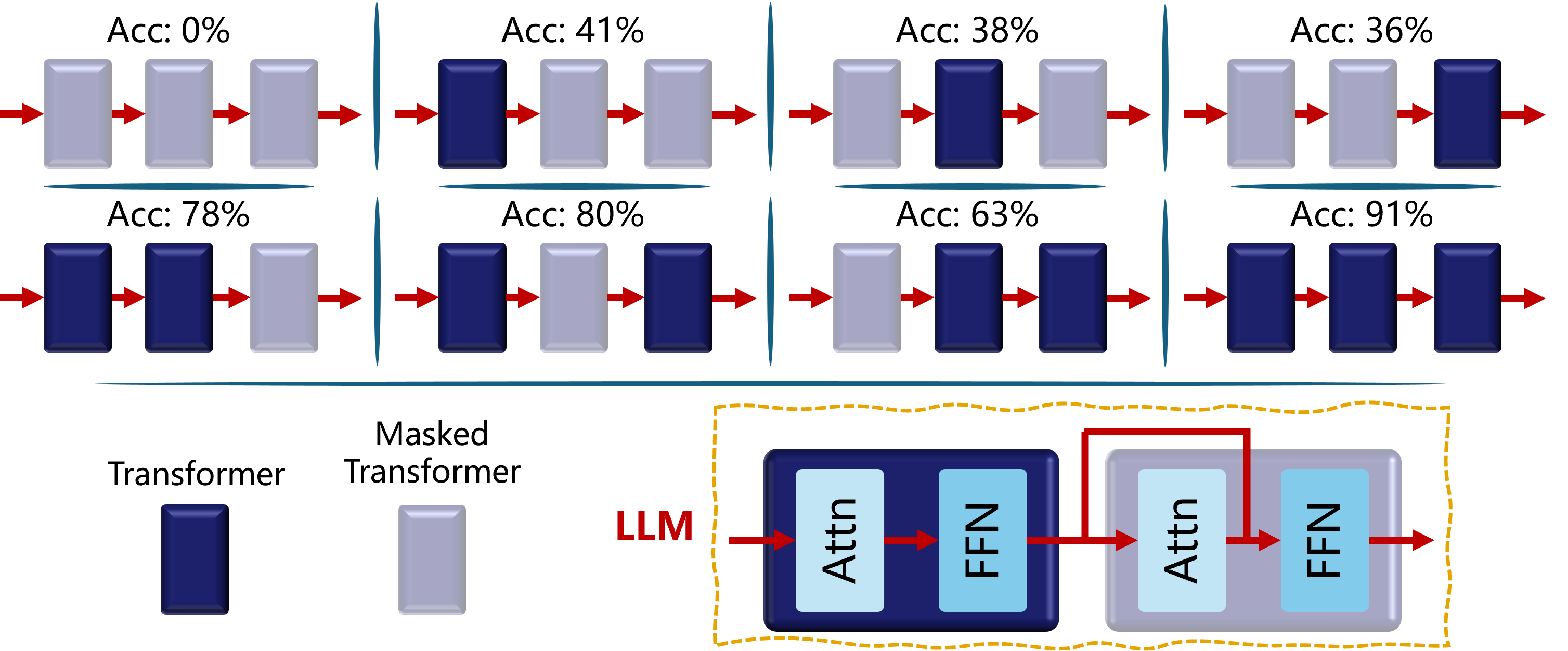}
\caption{Contribution analysis of 3 Transformer layers in an LLM. For intuitive presentation, we use accuracy to represent the contribution of different Transformer layers.}
\label{fig:shapley}
\vspace{-0.1in}
\end{wrapfigure}

SV is a unique contribution allocation scheme that satisfies a set of fairness axioms. Here, we leverage the concept to calculate the weighted average of all marginal contributions of each Transformer layer in an LLM. To illustrate the calculation process, we consider a simple LLM with 3 Transformer layers as shown in Figure \ref{fig:shapley}. Since all Transformer layers have the same structure, we can easily mask a Transformer layer during inference and do not affect SV calculation, as shown in the bottom right corner of Figure \ref{fig:shapley}. We use the accuracy of an LLM as the basis for calculating the SVs.

We first identify each Transformer layer’s contribution when they participate individually, when 2 participate together, and when all 3 participate together. Particularly, as an LLM without any Transformer layer cannot function normally, we assume the accuracy to be 0. Then, we consider all possible combinations of Transformer layers and calculate their marginal values (e.g. what value does each Transformer layer add when Transformer layer 1 enters the LLM first, followed by Transformer layer 2, and then Transformer layer 3). Finally, we need to add them up and work out the SV (i.e., the average) for each Transformer layer.

To formalize this process, an LLM consists of $\mathcal{T}$ Transformer layers and a value function $\nu$. Perplexity (PPL) is usually used to evaluate LLMs in LLM pruning \cite{jaiswalcompressing,xiasheared,ma2023llm}. The lower the value, the better the performance. Therefore, the value function is denoted as $\nu(\mathcal{T}) = 1/\text{PPL}(\mathcal{T})$. Particularly, if there is no Transformer layer, the value function must be 0, i.e., $\nu(\emptyset) = 0$. Let $\mathcal{S}\subseteq \mathcal{T}$ denote a subset of $\mathcal{T}$. According to \cite{shapley1953value}, the SV of any Transformer layer $t\in\mathcal{T}$ is:
\begin{equation}
\label{eq:shaply}
\phi_{t}=\sum_{\mathcal{S} \subseteq \mathcal{T}\setminus\{t\}} w_{\mathcal{S}}[\nu(\mathcal{S} \cup\{t\})-\nu(\mathcal{S})],\forall t\in\mathcal{T},
\end{equation}
where $w_{\mathcal{S}} = T \cdot\binom{T-1}{\mathcal{S}}=\frac{T!}{|\mathcal{S}|!(T-|\mathcal{S}|-1)!}$ is a coefficient. $\nu(\mathcal{S} \cup\{t\})-\nu(\mathcal{S})$ is known as Transformer layer $t$'s marginal contribution.
Eq. (\ref{eq:shaply}) calculaties the marginal contribution for every subset $\mathcal{S}$, which results in a combinatorial explosion as the number of Transformer layers increases. Specifically, for $T$ Transformer layers, the total number of possible subsets $\mathcal{S}$ is $2^T$. For instance, with 32 Transformer layers in LLaMa-7B, the number of possible subsets becomes $2^{32} = 4,294,967,296$. The exact computation of Shapley values requires evaluating every subset that excludes the given Transformer layer, meaning that $32 \times 2^{32}$ evaluations are needed. Each additional feature doubles the number of subsets that must be considered, causing an exponential increase in the number of computations. Consequently, this problem is NP-hard.

\subsection{Shapley Value Approximation}

\begin{wrapfigure}{R}{0.45\textwidth}
\vspace{-0.2in}
\centering
\includegraphics[width=1\linewidth]{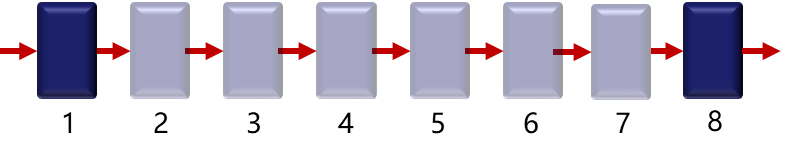}
\caption{Selection of Transformer coalitions under an LLM with 8 Transformer layers.}
\label{fig:transformer_sv}
\vspace{-0.1in}
\end{wrapfigure}

To address the NP-hard problem, this subsection proposes a SWSV method to estimate the contribution of Transformer layers. The key is that many coalitions of Transformer layers are useless and meaningless when evaluating LLMs. We consider an LLM with 8 Transformer layers as shown in Figure \ref{fig:transformer_sv}, where Transformers 1 and 8 are active and Transformer layers 2-7 are masked. We have two limitations: \textit{1) too few active Transformer layers; and 2) too far apart between active Transformer layers}.

Specifically, a well-trained LLM relies on maintaining a sufficiently large network structure, even in the presence of sparsity, to preserve its representation capacity and functionality \cite{li2022parameter,lee2024cats}. Sparsity, while effective in reducing computational costs, must be implemented judiciously to ensure that the structural integrity and critical pathways of the model remain intact. An LLM with an insufficient number of active Transformer layers—such as only two active layers (e.g., Transformer layers 1 and 8)—is highly likely to suffer from significant performance degradation. This occurs because such configurations disrupt the hierarchical processing of information, resulting in incomplete or suboptimal feature representations. Moreover, when many intermediate Transformer layers are masked, the direct transmission of outputs from an earlier Transformer layer (e.g., Transformer layer 1) to a much later one (e.g., Transformer layer 8) bypasses the essential intermediate processing steps. This shortcut diminishes the model's ability to learn and propagate nuanced information through the intermediate layers, leading to a distorted representation of the Shapley value. Such scenarios are problematic as they fail to reflect the contributions of individual layers within the model architecture.

To address these issues, it is critical to design an SV approximation method to avoid extreme patterns with few Transformer layers. The goal is to improve the accuracy of SV estimation, enhance model performance, and reduce computational complexity by preserving critical Transformer layers.

\begin{wrapfigure}{R}{0.45\textwidth}
\vspace{-0.2in}
\centering
\includegraphics[width=1\linewidth]{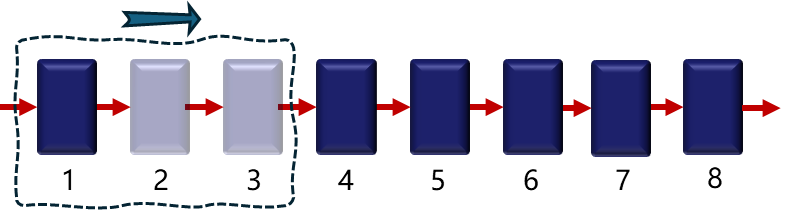}
\caption{Illustration of SWSV.}
\label{fig:swsva}
\vspace{-0.1in}
\end{wrapfigure}

To this end, we propose a sliding window-based Shapley value approximation method for evaluating the contribution of Transformer layers in LLMs (Figure \ref{fig:swsva}). When calculating the SV of Transformer layer $t\in\mathcal{T}$, we consider the sliding window with $N=|\mathcal{S}_t|$ Transformer layers, where these Transformer layers are closely connected. Figure \ref{fig:swsva} shows the sliding window with 3 Transformer layers, where the SV of Transformer layer 1 is only related to the Transformer layers in this sliding window. It is expressed as:
\begin{equation}
\label{eq:shaply_apx}
\hat{\phi}_{t}=\sum_{\mathcal{S} \subseteq \mathcal{S}_t\setminus\{t\}} w_{\mathcal{S}}[\nu(\mathcal{S}\cup\mathcal{T}_t \cup\{t\})-\nu(\mathcal{S}\cup\mathcal{T}_t)],\forall t\in\mathcal{T},
\end{equation}
where $\mathcal{T}_t=\mathcal{T}\setminus\mathcal{S}_t$ is coalitions without $\mathcal{S}_t$.

\begin{wrapfigure}{r}{0.46\textwidth}
\vspace{-0.2in}
\hfill
\begin{minipage}{0.44\textwidth}
\begin{algorithm}[H]
\label{alg:sva}
\caption{SWSV}
\KwIn{$\mathcal{T}$, $N$, $\mathcal{V}$, and a dataset\;}
\KwOut{Shapley values $\{\hat{\phi}_{t}\}_{t\in\mathcal{T}}$\;}
\While{$t\in\mathcal{T}$}{
    Determine the sliding window $\mathcal{S}_t$\;
    \For{$\mathcal{S}$ in all coalitions of $\mathcal{S}_t\setminus\{t\}$}{
        \If{$\mathcal{S}\cup\mathcal{T}_t \cup\{t\} \not\subseteq \mathcal{V}$}{
            Put $\mathcal{S}\cup\mathcal{T}_t \cup\{t\}$ into $\mathcal{V}$\;
            Calculate $\nu(\mathcal{S}\cup\mathcal{T}_t \cup\{t\})$\;
        }
        \If{$\mathcal{S}\cup\mathcal{T}_t \not\subseteq \mathcal{V}$}{
            Put $\mathcal{S}\cup\mathcal{T}_t$ into $\mathcal{V}$\;
            Calculate $\nu(\mathcal{S}\cup\mathcal{T}_t)$\;
        }
    }
    Calculate $\hat{\phi}_{t}$ by (\ref{eq:shaply_apx})\;
}
\end{algorithm}
\end{minipage}
\vspace{-0.2in}
\end{wrapfigure}

Neighboring Transformer layers around a target Transformer layer $t$ play a significant role in determining its SV, as they strongly influence the context and interactions that contribute to the target Transformer layer's functionality. To effectively capture these interdependencies, a sliding window strategy is employed, which selects a set of neighboring Transformer layers both before and after the target Transformer layer. This approach ensures that the SV calculation takes into account the most relevant local interactions within the Transformer layer stack. The sliding window size $N$ is generally set to an odd number, such as 3, 5, or 7, to maintain symmetry around the target Transformer layer. For instance, with a sliding window of size 3, the calculation includes one preceding Transformer layer ($t-1$), the target Transformer layer itself ($t$), and one succeeding Transformer layer ($t+1$). This balanced selection ensures that contributions from both upstream and downstream Transformer layers are equally considered, facilitating a more accurate evaluation of the target Transformer layer's importance.

\begin{wrapfigure}{r}{0.42\textwidth}
\vspace{-0.2in}
\hfill
\begin{minipage}{0.4\textwidth}
\begin{algorithm}[H]
\label{alg:pruning}
\caption{\methodname{}}
\KwIn{an LLM and a dataset\;}
\KwOut{The Pruned LLM\;}
Calculate the contributions $\{\hat{\phi}_{t}\}_{t\in\mathcal{T}}$ of Transformer layers by SWSV\;
Allocate the pruning ratios $\{\rho_t\}_{t\in\mathcal{T}}$ by Eq. (\ref{eq:ratio})\;
\While{$t\in\mathcal{T}$}{
    Prune the Transformer layer $t$ by \textbf{an advanced pruning method} at the pruning ratio $\rho_t$\;
}
Evaluate the pruned LLM\;
\end{algorithm}
\end{minipage}
\vspace{-0.2in}
\end{wrapfigure}

Special considerations are made for boundary Transformer layers, such as the first and last ones, where the default sliding window size might result in insufficient neighbors. In these cases, the sliding window is adjusted to include the available Transformer layers, while maintaining the specified number of neighbors as much as possible. For example, when evaluating the first Transformer layer, additional downstream Transformer layers can be included to compensate for the absence of preceding Transformer layers. Similarly, for the last Transformer layer, upstream Transformer layers are prioritized. This ensures consistency in the Shapley value calculations across all Transformer layers, preserving the intended neighborhood size while respecting the structural boundaries of the model. The pseudocode of SWSV is shown in Algorithm \ref{alg:sva}. Its computational complexity is $\mathcal{O}(T2^{N-1})$, significantly lower than the original $\mathcal{O}(2^{T})$. For instance, computing contributions for LLaMA-7B with $N=5$ requires only 512 subsets, compared to $4,294,967,296$ in the original approach.

\subsection{\methodname{} for LLMs}
After calculating the contributions of Transformer layers (${\hat{\phi}_{t}},{t\in\mathcal{T}}$), these contributions can be utilized to allocate the pruning ratio for each Transformer layer. Transformer layers with greater contributions are considered more important and thus have smaller pruning ratios. Therefore, the pruning ratio $\rho_t$ is inversely proportional to the contribution, i.e., $\rho_t \propto 1 / \hat{\phi}_{t}$. However, since overly sparse Transformer layers can degrade the performance of LLMs, it is necessary to constrain the allocated pruning ratios. To achieve this, we introduce a small positive value $\lambda$ (e.g., 0.05 or 0.1) as a boundary. The allocated pruning ratios are restricted to lie within the given pruning ratio range, with an allowable error of $\lambda$, i.e., $\rho_t \in [\rho - \lambda, \rho + \lambda]$. The detailed mathematical expression is:
\begin{equation}
\label{eq:ratio}
\rho_t = \rho-a_t+\text{mean}\{a_t\}_{t\in\mathcal{T}}, 
\end{equation}
where $a_t=\frac{2\lambda(\hat{\phi}_{t}-\min{\{\hat{\phi}_{t}}\}_{t\in\mathcal{T}})}{\max{\{\hat{\phi}_{t}}\}_{t\in\mathcal{T}}-\min{\{\hat{\phi}_{t}}\}_{t\in\mathcal{T}}}$.

\methodname{} is a framework which can be applied to advanced pruning methods (e.g., Wanda \cite{sun2023simple} and SparseGPT \cite{frantar2023sparsegpt}). The pseudocode of \methodname{} is presented in Algorithm \ref{alg:pruning}.

\section{Experimental Evalution}
\label{sec:exper}

\textbf{LLMs and Baselines.} We evaluate \methodname{} on the LLaMA-V1/v2/v3, and OPT model families \cite{frantar2023sparsegpt}. Due to computational constraints, we select LLMs with up to 13 billion parameters, quantized to float16 with a sequence length of 2048. We select 3 baselines for comparison: 1) Magnitude (2017) \cite{zhu2017prune}, 2) Wanda (2024) \cite{sun2023simple}, and 3) SparseGPT (2023) \cite{frantar2023sparsegpt}. The proposed \methodname{} method can be integrated into these baselines, allowing us to directly assess its performance gains. In addition, we compare \methodname{} with OWL (2024) \cite{yin2023outlier} and ALS (2024) \cite{li2024adaptive}, two pruning ratio allocation frameworks based on outliers and information orthogonality, respectively. We implement all experiments in PyTorch and use the HuggingFace library to download the pre-trained LLMs and datasets. All experiments are conducted on a server equipped with 4 NVIDIA A100 GPUs (40GB each), one AMD EPYC CPU, and 252 GB of memory. The pruning ratio constraint $\lambda$ is set to 0.1, and the sliding window size $N$ is chosen as 3, 5, or 7. \textit{Notably, since ALS does not publicly release its source code, we report its results directly from the original paper.}

\textbf{Experiment Settings.} We implement all experiments in PyTorch and use the HuggingFace library to download the pre-trained LLMs and datasets. All experiments are performed on a server with 4 NVIDIA A100 GPUs, 1 AMD EPYC CPU, and 252 GB of memory. To assess the performance of the pruned LLMs, we employ two general metrics: PPL and zero-shot evaluation \cite{sun2023simple}. PPL reflects the model's ability to predict the next word given the preceding context \cite{muhlgay2023generating}, where lower values indicate better performance. We evaluate PPL using the WikiText-2 dataset \cite{merity2016pointer}, selecting two randomly downloaded subsets of WikiText-2 for testing. For zero-shot evaluation, we choose 4 tasks—BoolQ, RTE, WinoGrande (WG), and OpenBookQA (OBQA)—from the EleutherAI LM Harness \cite{mihaylov2018can}, where higher scores indicate better performance. Detailed settings are shown in Appendix \ref{sec:setting}.

\begin{table}[t]
  \centering
  \caption{PPL\textdownarrow~ performance of LLaMA-v1/v2 family pruned by 3 existing methods with (w.) \methodname{} at 70\% sparsity. We highlight the improved performance in \color{blue}{\textbf{blue}}.}
  \resizebox*{1 \linewidth}{!}{
    \begin{tabular}{lcccc}
    \toprule
    Model & LLaMA-7B & LLaMA-13B & LLaMA2-7B & LLaMA2-13B \\
    \midrule
    Dense & 5.68  & 5.09  & 5.47  & 4.88  \\
    \midrule
    Magnitude & 48426.42  & 84531.15  & 49808.59  & 214.21  \\
    \rowcolor[rgb]{ .949,  .949,  .949} Magnitude w. \methodname{} & 30586.90 (\textbf{\color{blue}{-36.84\%}}) & 10632.55 (\textbf{\color{blue}{-87.42\%}}) & 31729.74 (\textbf{\color{blue}{-18.01\%}}) & 82.98 (\textbf{\color{blue}{-18.01\%}}) \\
    Wanda & 70.84  & 41.10  & 57.20  & 34.06  \\
    \rowcolor[rgb]{ .949,  .949,  .949} Wanda w. \methodname{} & 34.95 (\textbf{\color{blue}{-50.66\%}}) & 15.92 (\textbf{\color{blue}{-61.28\%}}) & 29.56 (\textbf{\color{blue}{-18.01\%}}) & 20.46 (\textbf{\color{blue}{-18.01\%}}) \\
    SparseGPT & 18.42  & 13.74  & 17.87  & 14.39  \\
    \rowcolor[rgb]{ .949,  .949,  .949} SparseGPT w. \methodname{} & 15.10 (\textbf{\color{blue}{-18.01\%}}) & 11.06 (\textbf{\color{blue}{-19.55\%}}) & 14.84 (\textbf{\color{blue}{-18.01\%}}) & 12.48 (\textbf{\color{blue}{-18.01\%}}) \\
    \bottomrule
    \end{tabular}}
  \label{tab:high_sparsity}
\vspace{-0.1in}
\end{table}%

\textbf{PPL Improvement by \methodname{}.} We first evaluate the PPL performance gains achieved by integrating \methodname{} into LLM pruning methods, as shown in Table \ref{tab:high_sparsity}. \methodname{} significantly enhances existing pruning approaches, particularly Magnitude and Wanda, while SparseGPT demonstrates minimal improvements due to its already strong baseline performance. \methodname{} has 61.28 \% improvement on LLaMA-13B by Wanda. Wanda consistently benefits from \methodname{} across all LLaMA models, with the largest impact observed in larger models like LLaMA-13B, where \methodname{} effectively mitigates performance degradation. SparseGPT, despite its high baseline effectiveness, achieves further refinement with \methodname{}, highlighting the compatibility and synergy between the two. Overall, \methodname{} emerges as a crucial enhancement, stabilizing pruning outcomes and delivering superior results, especially when combined with advanced methods like SparseGPT.

\textbf{PPL and Zero-shot Performance.} 
The results in Table \ref{tab:ppl} underscore the effectiveness of integrating \methodname{}, OWL, and ALS into existing pruning methods for the LLaMA and OPT model families at 50\% sparsity. \textit{Particularly, as ALS does not release its source code, the results of ALS are not fair and are for reference only.} In most scenarios, \methodname{} consistently demonstrates superior performance. Specifically, Magnitude combined with \methodname{} consistently outperforms both its standalone version and Magnitude with OWL. Notably, while \methodname{} slightly improves the performance of Wanda and SparseGPT, the gains are less pronounced, as the pruned LLMs at 50\% sparsity already exhibit robust performance, leaving limited room for further enhancements. Interestingly, \methodname{}'s effectiveness is more apparent with larger LLMs, such as the 7-billion-parameter models, whereas its impact diminishes for smaller OPT-2.7B. 

\begin{table}[htbp]
  \centering
  \caption{PPL\textdownarrow~ performance of LLaMA-v1/v2/v3 family and OPT family pruned by different methods at 50\% sparsity. We highlight the best performance in \textbf{bold}. We set the results of ALS as \color{gray}{gray}.}
  \resizebox*{1\linewidth}{!}{
    \begin{tabular}{lcccccc}
    \toprule
    Model & LLaMA-7B & Vicuna-7B & LLaMA2-7B & LLaMA3.2-3B & OPT-2.7B & OPT-6.7B \\
    \midrule
    Dense & 5.6772  & 6.9031  & 5.4721  & 7.8137  & 12.4705  & 10.8602  \\
    \midrule
    Magnitude & 17.2882  & 24.0034  & 16.0301  & 139.4124  & 265.2033  & 968.7209  \\
    Magnitude w. ALS & \color{gray}{16.8000}  & --     & \color{gray}{\textbf{15.1900}} & --     & --     & \color{gray}{950.0000}  \\
    Magnitude w. OWL & 16.3453  & 21.3578  & 15.7421  & 99.2163  & 207.0598  & 363.6955  \\
    \rowcolor[rgb]{ .949,  .949,  .949} Magnitude w. \methodname{} & \textbf{15.9755 } & \textbf{20.7409 } & 15.3144  & \textbf{89.4710 } & \textbf{158.2674 } & \textbf{308.4234 } \\
    Wanda & 7.0884  & 8.5325  & \textbf{6.7741 } & 12.6977  & \textbf{13.9611 } & 12.0780  \\
    Wanda w. ALS & \color{gray}{12.4700}  & --     & \color{gray}{11.6100}  & --     & --     & \color{gray}{19.1600}  \\
    Wanda w. OWL & 7.0941  & 8.5731  & 6.7954  & 12.6735  & 14.7622  & 12.3921  \\
    \rowcolor[rgb]{ .949,  .949,  .949} Wanda w. \methodname{} & \textbf{7.0610 } & \textbf{8.4428 } & 6.7959  & \textbf{12.1578 } & 14.0323  & \textbf{12.0065 } \\
    SparseGPT & 6.8701  & 8.0767  & 6.6000  & 11.0357  & \textbf{12.0591 } & 11.2711  \\
    SparseGPT w. ALS & \color{gray}{11.8700}  & --     & \color{gray}{10.9900}  & --     & --     & \color{gray}{12.2900}  \\
    SparseGPT w. OWL & 6.8785  & 8.2247  & 6.6388  & 10.9828  & 13.3490  & 11.5101  \\
    \rowcolor[rgb]{ .949,  .949,  .949} SparseGPT w. \methodname{} & \textbf{6.8336 } & \textbf{8.0762 } & \textbf{6.5843 } & \textbf{10.8495 } & 12.9768  & \textbf{11.2582 } \\
    \bottomrule
    \end{tabular}}
  \label{tab:ppl}%
\vspace{-0.1in}
\end{table}%

Table \ref{tab:sparsity} compares PPL of pruning methods (Magnitude, Wanda, SparseGPT) combined with ALS, OWL, and \methodname{} across LLaMA-7B, LLaMA2-7B, and OPT-6.7B at 30–50\% sparsity. \methodname{} achieves the lowest PPL in nearly all cases, demonstrating consistent superiority. For example, on OPT-6.7B at 50\% sparsity, \methodname{} attains a PPL of 308.4234 (Magnitude) and 12.0065 (Wanda), markedly better than OWL (363.6955, 12.3921) and ALS (950.0000, 19.1600). The gains are especially pronounced at higher sparsities and for larger models, highlighting \methodname{}’s robustness and scalability. These results solidify \methodname{} as a state-of-the-art pruning technique for efficient model compression.

\begin{table}[htbp]
  \centering
  \caption{PPL\textdownarrow~ performance of LLaMA-7B, LLaMA2-7B, and OPT-6.7B at different sparsities.}
  \resizebox*{1\linewidth}{!}{
    \begin{tabular}{l|ccc|ccc|ccc}
    \toprule
     Model  & \multicolumn{3}{c|}{LLaMA-7B} & \multicolumn{3}{c|}{LLaMA2-7B} & \multicolumn{3}{c}{OPT-6.7B} \\
    \midrule
    Sparsity & 30\%  & 40\%  & 50\%  & 30\%  & 40\%  & 50\%  & 30\%  & 40\%  & 50\% \\
    \midrule
    Magnitude w. ALS & --     & --     & \textcolor[rgb]{ .647,  .647,  .647}{16.8000 } & \textcolor[rgb]{ .647,  .647,  .647}{9.6000 } & \textcolor[rgb]{ .647,  .647,  .647}{11.0300 } & \textcolor[rgb]{ .647,  .647,  .647}{\textbf{15.1900 }} & --     & --     & \textcolor[rgb]{ .647,  .647,  .647}{950.0000 } \\
    Magnitude w. OWL & 6.7413  & 8.8245  & 16.3453  & 6.3403  & 8.1432  & 15.7421  & 12.8504  & 20.4357  & 363.6955  \\
    \rowcolor[rgb]{ .949,  .949,  .949} Magnitude w. \methodname{} & \textbf{6.6769 } & \textbf{8.5035 } & \textbf{15.9755 } & \textbf{6.3438 } & \textbf{8.2723 } & 15.3144  & \textbf{12.3753 } & \textbf{18.7027 } & \textbf{308.4234 } \\
    Wanda w. ALS & --     & --     & \textcolor[rgb]{ .647,  .647,  .647}{12.4700 } & \textcolor[rgb]{ .647,  .647,  .647}{9.1500 } & \textcolor[rgb]{ .647,  .647,  .647}{9.8100 } & \textcolor[rgb]{ .647,  .647,  .647}{11.6100 } & --     & --     & \textcolor[rgb]{ .647,  .647,  .647}{19.1600 } \\
    Wanda w. OWL & 6.0066  & 6.3583  & 7.0941  & 5.7712  & 6.0827  & \textbf{6.7954 } & 10.7365  & 11.2408  & 12.3921  \\
    \rowcolor[rgb]{ .949,  .949,  .949} Wanda w. \methodname{} & \textbf{5.9885 } & \textbf{6.3294 } & \textbf{7.0610 } & \textbf{5.7554 } & \textbf{6.0702 } & 6.7959  & \textbf{10.6790 } & \textbf{11.1051 } & \textbf{12.0065 } \\
    SparseGPT w. ALS & --     & --     & \textcolor[rgb]{ .647,  .647,  .647}{11.8700 } & \textcolor[rgb]{ .647,  .647,  .647}{9.1100 } & \textcolor[rgb]{ .647,  .647,  .647}{9.6700 } & \textcolor[rgb]{ .647,  .647,  .647}{10.9900 } & --     & --     & \textcolor[rgb]{ .647,  .647,  .647}{12.2900 } \\
    SparseGPT w. OWL & 5.9512  & 6.2543  & 6.8785  & 5.7599  & 6.0340  & 6.6388  & 10.9449  & 11.0805  & 11.5101  \\
    \rowcolor[rgb]{ .949,  .949,  .949} SparseGPT w. \methodname{} & \textbf{5.9483 } & \textbf{6.2285 } & \textbf{6.8336 } & \textbf{5.7535 } & \textbf{6.0306 } & \textbf{6.6243 } & \textbf{10.8853 } & \textbf{10.9279 } & \textbf{11.2582 } \\
    \bottomrule
    \end{tabular}}
  \label{tab:sparsity}
\vspace{-0.1in}
\end{table}%

\begin{table}[htbp]
  \centering
  \caption{Zero-shot performance\textuparrow~ of LLaMA-7B and LLaMA2-7B pruned by different methods at 50\% sparsity. The "mean" is the average of BoolQ, RTE, WG, and OBQA.}
  \resizebox*{1\linewidth}{!}{
    \begin{tabular}{l|ccccc|ccccc}
    \toprule
    Model & \multicolumn{5}{c|}{LLaMA-7B}         & \multicolumn{5}{c}{LLaMA2-7B} \\
    \midrule
    Zero-shot task & BoolQ & RTE   & WG    & OBQA  & Mean\textuparrow~  & BoolQ & RTE   & WG    & OBQA  & Mean\textuparrow~ \\
    \midrule
    Dense & 75.08\% & 66.06\% & 70.09\% & 34.20\% & 61.36\% & 77.71\% & 62.82\% & 69.14\% & 31.40\% & 60.27\% \\
    \midrule
    Magnitude & 54.56\% & 54.15\% & 59.43\% & 22.60\% & 47.68\% & 62.94\% & 57.04\% & 63.38\% & 26.80\% & 52.54\% \\
    Magnitude w. ALS & \textcolor[rgb]{ .647,  .647,  .647}{59.82\%} & \textcolor[rgb]{ .647,  .647,  .647}{54.51\%} & \textcolor[rgb]{ .647,  .647,  .647}{61.25\%} & \textcolor[rgb]{ .647,  .647,  .647}{36.60\%} & \textcolor[rgb]{ .647,  .647,  .647}{\textbf{53.05\%}} & \textcolor[rgb]{ .647,  .647,  .647}{71.38\%} & \textcolor[rgb]{ .647,  .647,  .647}{55.24\%} & \textcolor[rgb]{ .647,  .647,  .647}{65.19\%} & \textcolor[rgb]{ .647,  .647,  .647}{41.40\%} & \textcolor[rgb]{ .647,  .647,  .647}{\textbf{58.30\%}} \\
    Magnitude w. OWL & 58.69\% & 55.23\% & 61.56\% & 27.80\% & 50.82\% & 63.88\% & 53.79\% & 62.90\% & 28.60\% & 52.29\% \\
    \rowcolor[rgb]{ .949,  .949,  .949} Magnitude w. \methodname{} & 57.63\% & 57.76\% & 61.40\% & 27.20\% & \underline{51.00\%} & 64.92\% & 55.96\% & 64.64\% & 28.60\% & \underline{53.53\%} \\
    Wanda & 72.45\% & 55.40\% & 66.61\% & 28.60\% & 55.77\% & 75.81\% & 54.87\% & 68.35\% & 31.20\% & 57.56\% \\
    Wanda w. ALS & \textcolor[rgb]{ .647,  .647,  .647}{73.70\%} & \textcolor[rgb]{ .647,  .647,  .647}{60.65\%} & \textcolor[rgb]{ .647,  .647,  .647}{66.30\%} & \textcolor[rgb]{ .647,  .647,  .647}{38.60\%} & \textcolor[rgb]{ .647,  .647,  .647}{\textbf{59.81\%}} & \textcolor[rgb]{ .647,  .647,  .647}{75.47\%} & \textcolor[rgb]{ .647,  .647,  .647}{54.87\%} & \textcolor[rgb]{ .647,  .647,  .647}{67.80\%} & \textcolor[rgb]{ .647,  .647,  .647}{44.80\%} & \textcolor[rgb]{ .647,  .647,  .647}{\textbf{60.74\%}} \\
    Wanda w. OWL & 73.43\% & 53.43\% & 65.98\% & 30.60\% & 55.86\% & 77.22\% & 54.51\% & 68.51\% & 30.40\% & 57.66\% \\
    \rowcolor[rgb]{ .949,  .949,  .949} Wanda w. \methodname{} & 72.91\% & 55.23\% & 67.56\% & 29.80\% & \underline{56.38\%} & 76.18\% & 54.15\% & 69.40\% & 31.20\% & \underline{57.73\%} \\
    SparseGPT & 74.43\% & 49.82\% & 68.11\% & 28.20\% & 55.14\% & 70.46\% & 54.51\% & 67.01\% & 28.00\% & 55.00\% \\
    SparseGPT w. ALS & \textcolor[rgb]{ .647,  .647,  .647}{74.28\%} & \textcolor[rgb]{ .647,  .647,  .647}{54.87\%} & \textcolor[rgb]{ .647,  .647,  .647}{66.77\%} & \textcolor[rgb]{ .647,  .647,  .647}{39.00\%} & \textcolor[rgb]{ .647,  .647,  .647}{\textbf{58.73\%}} & \textcolor[rgb]{ .647,  .647,  .647}{70.98\%} & \textcolor[rgb]{ .647,  .647,  .647}{55.96\%} & \textcolor[rgb]{ .647,  .647,  .647}{67.96\%} & \textcolor[rgb]{ .647,  .647,  .647}{40.00\%} & \textcolor[rgb]{ .647,  .647,  .647}{\textbf{58.73\%}} \\
    SparseGPT w. OWL & 72.84\% & 54.51\% & 67.88\% & 26.00\% & 55.31\% & 71.50\% & 56.32\% & 68.19\% & 28.40\% & 56.10\% \\
    \rowcolor[rgb]{ .949,  .949,  .949} SparseGPT w. \methodname{} & 73.30\% & 54.51\% & 68.75\% & 26.80\% & \underline{55.84\%} & 68.84\% & 63.54\% & 67.96\% & 28.60\% & \underline{57.23\%} \\
    \bottomrule
    \end{tabular}}
  \label{tab:zero-shot}
\vspace{-0.1in}
\end{table}%

Table \ref{tab:zero-shot} presents the zero-shot performance of various pruning methods applied to LLaMA-7B and LLaMA2-7B across four downstream tasks. The unpruned dense models serve as baselines, achieving 61.36\% and 60.27\% average accuracy, respectively. Among the pruning strategies, magnitude pruning yields the most significant performance degradation, whereas both Wanda and SparseGPT show stronger resilience, with SparseGPT with \methodname{} achieving up to 57.23\% on LLaMA2-7B, closely matching the dense baseline. Notably, our proposed method \methodname{} consistently improves performance across all pruning strategies and model backbones. It is worth noting that the results of ALS occasionally outperform the dense models; we suspect this is due to differences in experimental settings or hyperparameters rather than true pruning benefits. \textit{Thus, ALS results are reported for reference only and are excluded from direct comparison.} Overall, the results validate the effectiveness and robustness of \methodname{} in mitigating performance loss introduced by non-uniform pruning.

\begin{wraptable}{r}{0.5\textwidth}
\vspace{-0.2in}
  \centering
  \caption{PPL\textdownarrow~ under different $N$ at 50\% sparsity.}
  \resizebox*{1\linewidth}{!}{
    \begin{tabular}{lccc}
    \toprule
    LLaMA-7B & $N=3$ & $N=5$ & $N=7$ \\
    \midrule
    Magnitude w. \methodname{} & 16.0491  & 16.0081  & \textbf{15.9755 } \\
    Wanda w. \methodname{} & 7.0932  & \textbf{7.0610 } & 7.0921  \\
    SparseGPT w. \methodname{} & 6.8382  & \textbf{6.8336 } & 6.8487  \\
    \midrule
    OPT-6.7B & $N=3$ & $N=5$ & $N=7$ \\
    \midrule
    Magnitude w. \methodname{} & 525.4074  & \textbf{308.4234 } & 661.8835  \\
    Wanda w. \methodname{} & 12.0469  & 12.0339  & \textbf{12.0065 } \\
    SparseGPT w. \methodname{} & 11.2766  & 11.2661  & \textbf{11.2582 } \\
    \bottomrule
    \end{tabular}}
  \label{tab:size}
\vspace{-0.1in}
\end{wraptable}

\textbf{Windows Size Selection.}
Table \ref{tab:size} reports PPL of pruned LLaMA-7B and OPT-6.7B using different pruning strategies with \methodname{}, evaluated under varying window sizes ($N{=}3,5,7$). Across both models and all pruning methods, the PPL remains largely stable as the window size changes, indicating that \methodname{} is robust to the choice of window size. These results suggest that \methodname{} does not require careful tuning of the window size.

\begin{wraptable}{r}{0.65\textwidth}
  \centering
  \caption{Pruning ratio allocation by two SV approximation methods on LLaMA-7B and OPT-6.7B at 50\% sparsity.}
  \resizebox*{1 \linewidth}{!}{
    \begin{tabular}{clcccccc}
    \toprule
          &       & PPL\textdownarrow~   & BoolQ & RTE   & WG    & OBQA  & Mean\textuparrow~ \\
    \midrule
    \multirow{6}[1]{*}{\begin{sideways}LLaMA-7B\end{sideways}} & \cellcolor[rgb]{ .949,  .949,  .949}\methodname{} w. Magnitude & \cellcolor[rgb]{ .949,  .949,  .949} \textbf{15.98 } & \cellcolor[rgb]{ .949,  .949,  .949} 0.58  & \cellcolor[rgb]{ .949,  .949,  .949} 0.58  & \cellcolor[rgb]{ .949,  .949,  .949} 0.61  & \cellcolor[rgb]{ .949,  .949,  .949} 0.27  & \cellcolor[rgb]{ .949,  .949,  .949} \textbf{0.5100 } \\
          & SV w. Magnitude & 17.48  & 0.56  & 0.57  & 0.61  & 0.25  & 0.4956  \\
          & \cellcolor[rgb]{ .949,  .949,  .949}\methodname{} w. Wanda & \cellcolor[rgb]{ .949,  .949,  .949} \textbf{7.06 } & \cellcolor[rgb]{ .949,  .949,  .949} 0.73  & \cellcolor[rgb]{ .949,  .949,  .949} 0.59  & \cellcolor[rgb]{ .949,  .949,  .949} 0.68  & \cellcolor[rgb]{ .949,  .949,  .949} 0.30  & \cellcolor[rgb]{ .949,  .949,  .949} \textbf{0.5638 } \\
          & SV w. Wanda & 7.11  & 0.73  & 0.55  & 0.67  & 0.30  & 0.5626  \\
          & \cellcolor[rgb]{ .949,  .949,  .949}\methodname{} w. SparseGPT & \cellcolor[rgb]{ .949,  .949,  .949} \textbf{6.83 } & \cellcolor[rgb]{ .949,  .949,  .949} 0.73  & \cellcolor[rgb]{ .949,  .949,  .949} 0.55  & \cellcolor[rgb]{ .949,  .949,  .949} 0.69  & \cellcolor[rgb]{ .949,  .949,  .949} 0.27  & \cellcolor[rgb]{ .949,  .949,  .949} \textbf{0.5584 } \\
          & SV w. SparseGPT & 6.86  & 0.72  & 0.52  & 0.68  & 0.28  & 0.5493  \\
    \midrule
    \multirow{6}[2]{*}{\begin{sideways}OPT-6.7B\end{sideways}} & \cellcolor[rgb]{ .949,  .949,  .949}\methodname{} w. Magnitude & \cellcolor[rgb]{ .949,  .949,  .949} \textbf{308.42 } & \cellcolor[rgb]{ .949,  .949,  .949} 0.39  & \cellcolor[rgb]{ .949,  .949,  .949} 0.53  & \cellcolor[rgb]{ .949,  .949,  .949} 0.56  & \cellcolor[rgb]{ .949,  .949,  .949} 0.21  & \cellcolor[rgb]{ .949,  .949,  .949} \textbf{0.4217 } \\
          & SV w. Magnitude & 660.89  & 0.38  & 0.53  & 0.53  & 0.20  & 0.4089  \\
          & \cellcolor[rgb]{ .949,  .949,  .949}\methodname{} w. Wanda & \cellcolor[rgb]{ .949,  .949,  .949} \textbf{12.01 } & \cellcolor[rgb]{ .949,  .949,  .949} 0.62  & \cellcolor[rgb]{ .949,  .949,  .949} 0.53  & \cellcolor[rgb]{ .949,  .949,  .949} 0.61  & \cellcolor[rgb]{ .949,  .949,  .949} 0.24  & \cellcolor[rgb]{ .949,  .949,  .949} \textbf{0.4993 } \\
          & SV w. Wanda & 12.06  & 0.62  & 0.53  & 0.61  & 0.23  & 0.4977  \\
          & \cellcolor[rgb]{ .949,  .949,  .949}\methodname{} w. SparseGPT & \cellcolor[rgb]{ .949,  .949,  .949} \textbf{11.26 } & \cellcolor[rgb]{ .949,  .949,  .949} 0.64  & \cellcolor[rgb]{ .949,  .949,  .949} 0.55  & \cellcolor[rgb]{ .949,  .949,  .949} 0.63  & \cellcolor[rgb]{ .949,  .949,  .949} 0.25  & \cellcolor[rgb]{ .949,  .949,  .949} \textbf{0.5208 } \\
          & SV w. SparseGPT & 11.33  & 0.63  & 0.53  & 0.63  & 0.25  & 0.5111  \\
    \bottomrule
    \end{tabular}
    }
  \label{tab:sv_approx}
\vspace{-0.1in}
\end{wraptable}

\textbf{Different SV Approximation Methods.}
Table \ref{tab:sv_approx} shows our proposed SWSV method with the existing SV approximation method in \cite{kolpaczki2024approximating} to demonstrate its efficiency. For LLaMA-7B, SWSV with SparseGPT achieves the highest mean score (0.5584), slightly surpassing SV with SparseGPT (0.5493), showcasing its ability to enhance performance. These results demonstrate that SWSV not only provides marginal performance improvements over standard SV, but also offers lower computational complexity.

\begin{wrapfigure}{R}{0.45\textwidth}
\vspace{-0.1in}
\centering
\includegraphics[width=1\linewidth]{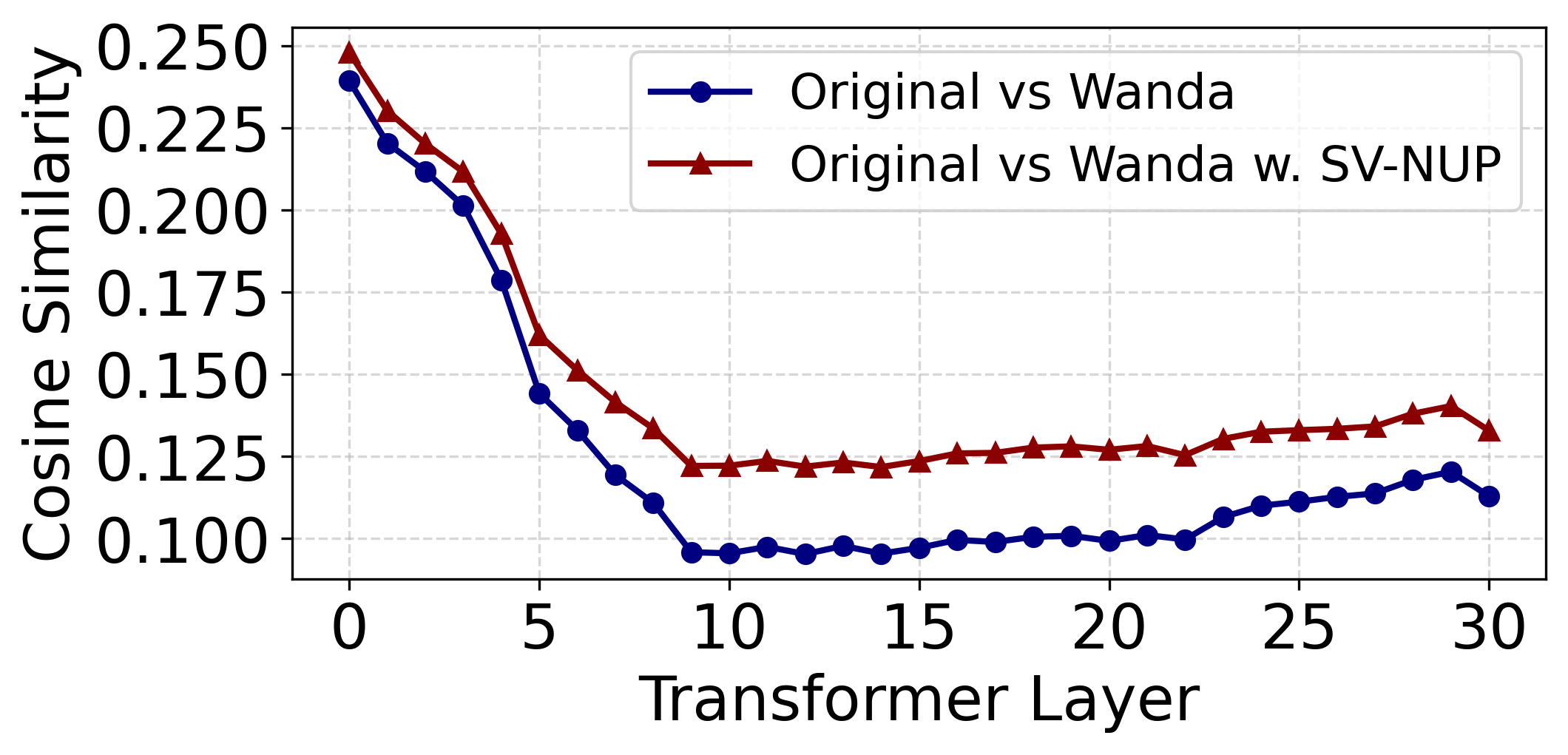}
\caption{Activation cosine similarity.}
\label{fig:similarity}
\vspace{-0.1in}
\end{wrapfigure}

\textbf{Activation Cosine Similarity.}
Figure \ref{fig:similarity} illustrates the activation cosine similarity across Transformer layers between the original LLaMA-7B and its pruned counterparts using Wanda and Wanda w. \methodname{} at 70\% sparsity. Notably, Wanda w. \methodname{} consistently yields higher similarity scores than Wanda alone across all layers (0–30), suggesting superior preservation of the original LLaMA-7B activations post-pruning. As pruning error increases, cosine similarity correspondingly declines. Further insights into \methodname{}’s advantages over other baselines are provided in Figure \ref{fig:ratio} in Appendix \ref{sec:results}. These findings substantiate the efficacy of SV-NUP in mitigating architectural distortion during pruning. Additional experimental results are presented in Appendix \ref{sec:results}.

\section{Conclusions and Future Work}
In this paper, we propose \methodname{}, a novel approach for non-uniform pruning of LLMs based on the contribution by each Transformer layer to overall model performance. By leveraging the Shapley value, we can assess the importance of individual Transformer layers within LLMs. To further address the computational complexity associated with the Shapley value, we design a sliding window-based approximation method. Extensive experiments have been carried out on LLMs, including LLaMA-v1/v2/v3 and OPT model families in comparison with 5 state-of-the-art pruning methods. \methodname{} achieves significant improvements in both PPL and zero-shot performance, demonstrating its promise as a useful LLM pruning method that can better preserve model performance.

In subsequent research, we plan to investigate the integration of pruning and quantization for compressing LLMs to reduce computation and storage.

\bibliographystyle{unsrt}
\bibliography{ref}

\appendix

\section{Extende Related Work}
\label{sec:related}

\textbf{Uniform Pruning.} Traditional pruning requires a round of re-training to restore performance, which poses significant challenges for LLMs. Researchers have developed pruning algorithms specifically tailored for LLM compression. For instance, \cite{ma2023llm} investigated structured sparse LLMs by applying Taylor pruning to remove entire weight rows, followed by LoRA fine-tuning \cite{hu2021lora}. In recent years, the focus has shifted toward unstructured pruning which eliminates the need for fine-tuning. SparseGPT \cite{frantar2023sparsegpt} employs the Hessian inverse for pruning, followed by weight updates to reduce reconstruction errors between dense and sparse weights. Wanda \cite{sun2023simple} introduced a criterion that incorporates weight magnitude and input activations to preserve outlier features.

\textbf{Non-uniform Pruning.} Uniform layerwise sparsity is commonly used for pruning language models \cite{zhu2017prune,gale2019state}, with several studies demonstrating its effectiveness in LLM pruning \cite{sanh2020movement,kurtic2022optimal}. However, there is a growing body of work exploring non-uniform layerwise sparsity, primarily in the context of vision models. For example, \cite{mocanu2016topological} proposed a non-uniform, scale-free topology inspired by graph theory, which outperforms dense counterparts when applied to restricted Boltzmann machines. Subsequent work has improved the scalability of this approach by leveraging Erdős–Rényi graphs \cite{erdds1959random}, extending the method to fully connected layers \cite{mocanu2018scalable} and convolutional layers \cite{evci2020rigging,liu2022unreasonable} to achieve data-free and feedforward-free layerwise sparsity. Another approach to non-uniform sparsity involves applying a global threshold across all layers \cite{frankle2018lottery,lee2018snip,wang2020picking,liu2021sparse}. However, global pruning has been found to be computationally expensive and ineffective when applied to LLMs.

\textbf{Analyzing LLMs.} The authors in \cite{shim2022understanding} analyzed the contributions of various components in LLMs and their impact on overall performance. \cite{gromov2024unreasonable} explored the role of deep layers in LLMs through layer pruning, providing insights into how different layers in relation to model performance. \cite{michel2019sixteen} examined the redundancy of attention heads in transformer-based models, demonstrating that many attention heads can be pruned without significant performance degradation. \cite{clark2019does} investigated the behavior of individual attention heads in BERT, revealing that each head serves a distinct role in capturing different linguistic features. Probing techniques are widely used to analyze the internal representations of LLMs. For instance, \cite{tenney2019bert} employed probing tasks to examine the linguistic information captured by BERT, finding that different layers encode distinct types of linguistic features. Furthermore, \cite{tenney2019you} introduced a suite of probes to analyze the representations learned by contextualized word embeddings, offering insights into how syntactic and semantic information is distributed across layers. Recently, \cite{zhang2024investigating} used Shapley values to evaluate the importance of layers in LLMs, providing a faithful assessment of their contributions. However, these studies have not explored SV-based non-uniform pruning of LLMs. \methodname{} bridges this gap.

\section{Implementation Details}
\label{sec:setting}
We follow existing works \cite{sunsimple, frantar2023sparsegpt, baisparsellm} and prune all linear layers in the FFN and MHA modules of LLMs. For calibration data, we use WikiText-2-v1, specifically "train-00000-of-00001.parquet" and "test-00000-of-00001.parquet" (\url{https://huggingface.co/datasets/Salesforce/wikitext/tree/main/wikitext-2-v1}), randomly selecting 32 segments of 2028 tokens from the trainset to ensure zero-shot pruning with generic internet text. For zero-shot evaluation, we adopt 4 tasks—BoolQ, RTE, WinoGrande (WG), and OpenBookQA (OBQA)—from the EleutherAI LM Harness framework modified by \cite{sunsimple}, enabling robust evaluation of pruned LLMs.

For \methodname{}, we use 32 segments of 2028 tokens from the WikiText-2-v1 trainset to compute Transformer layer contributions, with the sliding window size $N\in\{3,5,7\}$

For the SV approximation method \cite{kolpaczki2024approximating}, the budget and step size are set to $5\times10^3$ and 100, respectively.

For OWL \cite{yin2023outlier}, the outlier parameter is 5, and both OWL and \methodname{} share the same $\lambda$.

For ALS \cite{li2024adaptive}, since the original implementation has not been open-sourced, we directly report the results from the original paper as a reference. However, due to potential differences in experimental settings and the lack of reproducibility, these results are not strictly comparable to ours and should be considered as indicative rather than fully fair baselines.

\section{More Experimental Results}
\label{sec:results}

\begin{figure}[h]
\begin{minipage}[t]{0.45\textwidth}
\subfigure[LLaMA-7B with Mean] {\label{fig:ratio-a}
\includegraphics[width=0.7\linewidth]{./figs/magnitudellama7b_mean.png}
}

\subfigure[LLaMA-7B with Std] {\label{fig:ratio-b}
\includegraphics[width=0.7\linewidth]{./figs/magnitudellama7b_std.png}
}
\end{minipage}
\begin{minipage}[t]{0.45\textwidth}
\subfigure[Transformer Layer] {\label{fig:ratio-c}
\includegraphics[width=1\linewidth]{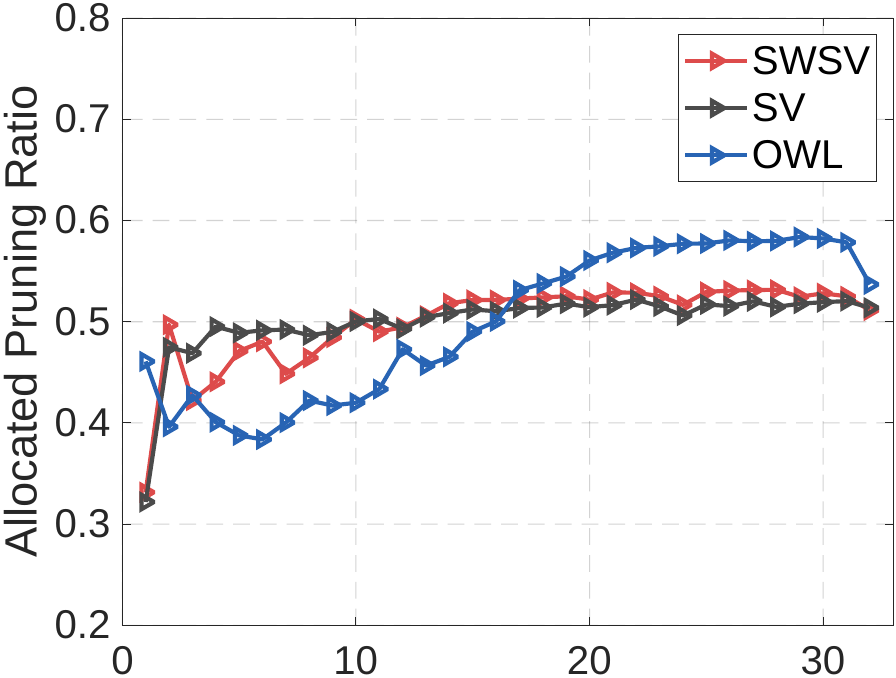}
}
\end{minipage}
\caption{Magnitude distribution of LLaMA-7B and pruning rate distribution allocated by the 3 methods at 50\% sparsity.}
\label{fig:ratio}
\end{figure}

\textbf{Analysis of Layer-wise Pruning Ratios.} Figure \ref{fig:ratio} visualizes the allocated pruning ratios across Transformer layers for LLaMA-7B (50\% sparsity) under SWSV, SV, and OWL. SWSV exhibits greater variability in early layers (e.g., Layer 5–15), with pruning ratios fluctuating ±15\% around the target sparsity before stabilizing in later layers. This contrasts with OWL’s near-uniform distribution and SV’s intermediate behavior. From the magnitude distribution in Figures \ref{fig:ratio-a} and \ref{fig:ratio-b}, SWSV’s dynamic allocation better aligns with layer sensitivity trends—aggressively pruning robust early layers while preserving critical later ones.

\begin{table}[htbp]
  \centering
  \caption{PPL\textdownarrow~ and Zero-shot\textuparrow~ performance of LLaMA-v1 pruned by different methods at 50\% sparsity.}
  \resizebox*{1\linewidth}{!}{
    \begin{tabular}{clcccccc}
    \toprule
          &       & PPL\textdownarrow~   & BoolQ & RTE   & WG    & OBQA  & Mean\textuparrow~ \\
    \midrule
    \multirow{10}[4]{*}{\begin{sideways}LLaMA-7B\end{sideways}} & Dense & 5.6772  & 75.08\% & 66.06\% & 70.09\% & 34.20\% & 61.36\% \\
\cmidrule{2-8}          & Magnitude & 17.2882  & 54.56\% & 54.15\% & 59.43\% & 22.60\% & 47.68\% \\
          & Magnitude w. OWL & 16.3453  & 58.69\% & 55.23\% & 61.56\% & 27.80\% & 50.82\% \\
          & \cellcolor[rgb]{ .949,  .949,  .949} Magnitude w. \methodname{} & \cellcolor[rgb]{ .949,  .949,  .949} \textbf{15.9755 } & \cellcolor[rgb]{ .949,  .949,  .949} 57.63\% & \cellcolor[rgb]{ .949,  .949,  .949} 57.76\% & \cellcolor[rgb]{ .949,  .949,  .949} 61.40\% & \cellcolor[rgb]{ .949,  .949,  .949} 27.20\% & \cellcolor[rgb]{ .949,  .949,  .949} \textbf{51.00\%} \\
          & Wanda & 7.0884  & 72.45\% & 55.40\% & 66.61\% & 28.60\% & 55.77\% \\
          & Wanda w. OWL & 7.0941  & 73.43\% & 53.43\% & 65.98\% & 30.60\% & 55.86\% \\
          & \cellcolor[rgb]{ .949,  .949,  .949} Wanda w. \methodname{} & \cellcolor[rgb]{ .949,  .949,  .949} \textbf{7.0610 } & \cellcolor[rgb]{ .949,  .949,  .949} 72.91\% & \cellcolor[rgb]{ .949,  .949,  .949} 55.23\% & \cellcolor[rgb]{ .949,  .949,  .949} 67.56\% & \cellcolor[rgb]{ .949,  .949,  .949} 29.80\% & \cellcolor[rgb]{ .949,  .949,  .949} \textbf{56.38\%} \\
          & SparseGPT & 6.8701  & 74.43\% & 49.82\% & 68.11\% & 28.20\% & 55.14\% \\
          & SparseGPT w. OWL & 6.8785  & 72.84\% & 54.51\% & 67.88\% & 26.00\% & 55.31\% \\
          & \cellcolor[rgb]{ .949,  .949,  .949} SparseGPT w. \methodname{} & \cellcolor[rgb]{ .949,  .949,  .949} \textbf{6.8336 } & \cellcolor[rgb]{ .949,  .949,  .949} 73.30\% & \cellcolor[rgb]{ .949,  .949,  .949} 54.51\% & \cellcolor[rgb]{ .949,  .949,  .949} 68.75\% & \cellcolor[rgb]{ .949,  .949,  .949} 26.80\% & \cellcolor[rgb]{ .949,  .949,  .949} \textbf{55.84\%} \\
    \midrule
    \multirow{10}[4]{*}{\begin{sideways}Vicuna-7B\end{sideways}} & Dense & 6.9031  & 78.10\% & 68.23\% & 69.38\% & 34.60\% & 62.58\% \\
\cmidrule{2-8}          & Magnitude & 24.0034  & 54.62\% & 52.71\% & 58.17\% & 22.00\% & 46.87\% \\
          & Magnitude w. OWL & 21.3578  & 60.67\% & 57.84\% & 60.56\% & 23.00\% & 50.52\% \\
          & \cellcolor[rgb]{ .949,  .949,  .949} Magnitude w. \methodname{} & \cellcolor[rgb]{ .949,  .949,  .949} \textbf{20.7409 } & \cellcolor[rgb]{ .949,  .949,  .949} 60.37\% & \cellcolor[rgb]{ .949,  .949,  .949} 58.84\% & \cellcolor[rgb]{ .949,  .949,  .949} 59.59\% & \cellcolor[rgb]{ .949,  .949,  .949} 23.60\% & \cellcolor[rgb]{ .949,  .949,  .949} \textbf{50.60\%} \\
          & Wanda & 8.5325  & 69.69\% & 68.59\% & 65.19\% & 29.60\% & 58.27\% \\
          & Wanda w. OWL & 8.5731  & 63.15\% & 70.40\% & 66.85\% & 29.60\% & 57.50\% \\
          & \cellcolor[rgb]{ .949,  .949,  .949} Wanda w. \methodname{} & \cellcolor[rgb]{ .949,  .949,  .949} \textbf{8.4428 } & \cellcolor[rgb]{ .949,  .949,  .949} 69.76\% & \cellcolor[rgb]{ .949,  .949,  .949} 69.68\% & \cellcolor[rgb]{ .949,  .949,  .949} 66.30\% & \cellcolor[rgb]{ .949,  .949,  .949} 28.80\% & \cellcolor[rgb]{ .949,  .949,  .949} \textbf{58.63\%} \\
          & SparseGPT & 8.0767  & 64.37\% & 65.34\% & 64.80\% & 29.40\% & 55.98\% \\
          & SparseGPT w. OWL & 8.2247  & 67.80\% & 67.15\% & 64.72\% & 28.40\% & 57.02\% \\
          & \cellcolor[rgb]{ .949,  .949,  .949} SparseGPT w. \methodname{} & \cellcolor[rgb]{ .949,  .949,  .949} \textbf{8.0762 } & \cellcolor[rgb]{ .949,  .949,  .949} 69.45\% & \cellcolor[rgb]{ .949,  .949,  .949} 68.23\% & \cellcolor[rgb]{ .949,  .949,  .949} 65.67\% & \cellcolor[rgb]{ .949,  .949,  .949} 30.20\% & \cellcolor[rgb]{ .949,  .949,  .949} \textbf{58.39\%} \\
    \bottomrule
    \end{tabular}}
  \label{tab:appdenix1}%
\end{table}%

\begin{table}[htbp]
  \centering
  \caption{PPL\textdownarrow~ and Zero-shot\textuparrow~ performance of LLaMA-v2/v3 pruned by different methods at 50\% sparsity.}
  \resizebox*{1\linewidth}{!}{
    \begin{tabular}{clcccccc}
    \toprule
          &       & PPL\textdownarrow~   & BoolQ & RTE   & WG    & OBQA  & Mean\textuparrow~ \\
    \midrule
    \multirow{10}[4]{*}{\begin{sideways}LLaMA2-7B\end{sideways}} & Dense & 5.4721  & 77.71\% & 62.82\% & 69.14\% & 31.40\% & 60.27\% \\
\cmidrule{2-8}          & Magnitude & 16.0301  & 62.94\% & 57.04\% & 63.38\% & 26.80\% & 52.54\% \\
          & Magnitude w. OWL & 15.7421  & 63.88\% & 53.79\% & 62.90\% & 28.60\% & 52.29\% \\
          & \cellcolor[rgb]{ .949,  .949,  .949} Magnitude w. \methodname{} & \cellcolor[rgb]{ .949,  .949,  .949} \textbf{15.3144 } & \cellcolor[rgb]{ .949,  .949,  .949} 64.92\% & \cellcolor[rgb]{ .949,  .949,  .949} 55.96\% & \cellcolor[rgb]{ .949,  .949,  .949} 64.64\% & \cellcolor[rgb]{ .949,  .949,  .949} 28.60\% & \cellcolor[rgb]{ .949,  .949,  .949} \textbf{53.53\%} \\
          & Wanda & \textbf{6.7741 } & 75.81\% & 54.87\% & 68.35\% & 31.20\% & 57.56\% \\
          & Wanda w. OWL & 6.7954  & 77.22\% & 54.51\% & 68.51\% & 30.40\% & 57.66\% \\
          & \cellcolor[rgb]{ .949,  .949,  .949} Wanda w. \methodname{} & \cellcolor[rgb]{ .949,  .949,  .949} 6.7959  & \cellcolor[rgb]{ .949,  .949,  .949} 76.18\% & \cellcolor[rgb]{ .949,  .949,  .949} 54.15\% & \cellcolor[rgb]{ .949,  .949,  .949} 69.40\% & \cellcolor[rgb]{ .949,  .949,  .949} 31.20\% & \cellcolor[rgb]{ .949,  .949,  .949} \textbf{57.73\%} \\
          & SparseGPT & 6.6000  & 70.46\% & 54.51\% & 67.01\% & 28.00\% & 55.00\% \\
          & SparseGPT w. OWL & 6.6388  & 71.50\% & 56.32\% & 68.19\% & 28.40\% & 56.10\% \\
          & \cellcolor[rgb]{ .949,  .949,  .949} SparseGPT w. \methodname{} & \cellcolor[rgb]{ .949,  .949,  .949} \textbf{6.5843 } & \cellcolor[rgb]{ .949,  .949,  .949} 68.84\% & \cellcolor[rgb]{ .949,  .949,  .949} 63.54\% & \cellcolor[rgb]{ .949,  .949,  .949} 67.96\% & \cellcolor[rgb]{ .949,  .949,  .949} 28.60\% & \cellcolor[rgb]{ .949,  .949,  .949} \textbf{57.23\%} \\
    \midrule
    \multirow{10}[4]{*}{\begin{sideways}LLaMA3.2-3B\end{sideways}} & Dense & 7.8137  & 73.27\% & 54.51\% & 69.77\% & 31.20\% & 57.19\% \\
\cmidrule{2-8}          & Magnitude & 139.4124  & 42.02\% & 53.43\% & 53.43\% & 14.20\% & 40.77\% \\
          & Magnitude w. OWL & 99.2163  & 41.28\% & 51.62\% & 54.78\% & 16.80\% & 41.12\% \\
          & \cellcolor[rgb]{ .949,  .949,  .949} Magnitude w. \methodname{} & \cellcolor[rgb]{ .949,  .949,  .949} \textbf{89.4710 } & \cellcolor[rgb]{ .949,  .949,  .949} 44.07\% & \cellcolor[rgb]{ .949,  .949,  .949} 51.99\% & \cellcolor[rgb]{ .949,  .949,  .949} 56.99\% & \cellcolor[rgb]{ .949,  .949,  .949} 17.60\% & \cellcolor[rgb]{ .949,  .949,  .949} \textbf{42.66\%} \\
          & Wanda & 12.6977  & 66.02\% & 55.23\% & 62.51\% & 25.60\% & 52.34\% \\
          & Wanda w. OWL & 12.6735  & 63.98\% & 53.07\% & 64.56\% & 24.80\% & 51.60\% \\
          & \cellcolor[rgb]{ .949,  .949,  .949} Wanda w. \methodname{} & \cellcolor[rgb]{ .949,  .949,  .949} \textbf{12.1578 } & \cellcolor[rgb]{ .949,  .949,  .949} 69.88\% & \cellcolor[rgb]{ .949,  .949,  .949} 55.96\% & \cellcolor[rgb]{ .949,  .949,  .949} 65.27\% & \cellcolor[rgb]{ .949,  .949,  .949} 24.40\% & \cellcolor[rgb]{ .949,  .949,  .949} \textbf{53.88\%} \\
          & SparseGPT & 11.0357  & 70.49\% & 52.71\% & 64.88\% & 24.40\% & 53.12\% \\
          & SparseGPT w. OWL & 10.9828  & 70.28\% & 49.46\% & 66.30\% & 24.80\% & 52.71\% \\
          & \cellcolor[rgb]{ .949,  .949,  .949} SparseGPT w. \methodname{} & \cellcolor[rgb]{ .949,  .949,  .949} \textbf{10.8495 } & \cellcolor[rgb]{ .949,  .949,  .949} 70.46\% & \cellcolor[rgb]{ .949,  .949,  .949} 51.62\% & \cellcolor[rgb]{ .949,  .949,  .949} 65.98\% & \cellcolor[rgb]{ .949,  .949,  .949} 25.20\% & \cellcolor[rgb]{ .949,  .949,  .949} \textbf{53.32\%} \\
    \bottomrule
    \end{tabular}}
  \label{tab:appdenix2}%
\end{table}%

\begin{table}[htbp]
  \centering
  \caption{PPL\textdownarrow~ and Zero-shot\textuparrow~ performance of OPT pruned by different methods at 50\% sparsity.}
  \resizebox*{1\linewidth}{!}{
    \begin{tabular}{clcccccc}
    \toprule
          &       & PPL\textdownarrow~   & BoolQ & RTE   & WG    & OBQA  & Mean\textuparrow~ \\
    \midrule
    \multirow{10}[4]{*}{\begin{sideways}OPT-2.7B\end{sideways}} & Dense & 12.4705  & 60.37\% & 54.87\% & 60.77\% & 25.00\% & 50.25\% \\
\cmidrule{2-8}          & Magnitude & 265.2033  & 39.66\% & 52.35\% & 53.43\% & 20.40\% & 41.46\% \\
          & Magnitude w. OWL & 207.0598  & 38.17\% & 52.71\% & 53.67\% & 21.20\% & 41.44\% \\
          & \cellcolor[rgb]{ .949,  .949,  .949} Magnitude w. \methodname{} & \cellcolor[rgb]{ .949,  .949,  .949} \textbf{158.2674 } & \cellcolor[rgb]{ .949,  .949,  .949} 38.59\% & \cellcolor[rgb]{ .949,  .949,  .949} 52.35\% & \cellcolor[rgb]{ .949,  .949,  .949} 53.51\% & \cellcolor[rgb]{ .949,  .949,  .949} 22.40\% & \cellcolor[rgb]{ .949,  .949,  .949} \textbf{41.71\%} \\
          & Wanda & \textbf{13.9611 } & 62.26\% & 51.99\% & 57.70\% & 20.40\% & 48.09\% \\
          & Wanda w. OWL & 14.7622  & 61.90\% & 52.35\% & 57.70\% & 20.80\% & 48.18\% \\
          & \cellcolor[rgb]{ .949,  .949,  .949} Wanda w. \methodname{} & \cellcolor[rgb]{ .949,  .949,  .949} 14.0042  & \cellcolor[rgb]{ .949,  .949,  .949} 62.29\% & \cellcolor[rgb]{ .949,  .949,  .949} 51.62\% & \cellcolor[rgb]{ .949,  .949,  .949} 58.41\% & \cellcolor[rgb]{ .949,  .949,  .949} 21.60\% & \cellcolor[rgb]{ .949,  .949,  .949} \textbf{48.48\%} \\
          & SparseGPT & 12.0591  & 60.70\% & 54.51\% & 61.33\% & 25.00\% & 50.39\% \\
          & SparseGPT w. OWL & 13.3490  & 62.87\% & 51.62\% & 58.80\% & 23.20\% & 49.12\% \\
          & \cellcolor[rgb]{ .949,  .949,  .949} SparseGPT w. \methodname{} & \cellcolor[rgb]{ .949,  .949,  .949} \textbf{12.9768 } & \cellcolor[rgb]{ .949,  .949,  .949} 63.30\% & \cellcolor[rgb]{ .949,  .949,  .949} 52.71\% & \cellcolor[rgb]{ .949,  .949,  .949} 59.43\% & \cellcolor[rgb]{ .949,  .949,  .949} 22.60\% & \cellcolor[rgb]{ .949,  .949,  .949} \textbf{49.51\%} \\
    \midrule
    \multirow{10}[4]{*}{\begin{sideways}OPT-6.7B\end{sideways}} & Dense & 10.8602  & 66.06\% & 55.23\% & 65.19\% & 27.60\% & 53.52\% \\
\cmidrule{2-8}          & Magnitude & 968.7209  & 38.04\% & 52.71\% & 50.59\% & 17.60\% & 39.74\% \\
          & Magnitude w. OWL & 363.6955  & 38.13\% & 52.71\% & 55.64\% & 21.40\% & 41.97\% \\
          & \cellcolor[rgb]{ .949,  .949,  .949} Magnitude w. \methodname{} & \cellcolor[rgb]{ .949,  .949,  .949} \textbf{308.4234 } & \cellcolor[rgb]{ .949,  .949,  .949} 39.08\% & \cellcolor[rgb]{ .949,  .949,  .949} 52.71\% & \cellcolor[rgb]{ .949,  .949,  .949} 55.67\% & \cellcolor[rgb]{ .949,  .949,  .949} 21.20\% & \cellcolor[rgb]{ .949,  .949,  .949} \textbf{42.17\%} \\
          & Wanda & 12.0780  & 62.14\% & 52.71\% & 60.14\% & 23.80\% & 49.70\% \\
          & Wanda w. OWL & 12.3921  & 62.20\% & 52.71\% & 58.64\% & 23.80\% & 49.34\% \\
          & \cellcolor[rgb]{ .949,  .949,  .949} Wanda w. \methodname{} & \cellcolor[rgb]{ .949,  .949,  .949} \textbf{12.0065 } & \cellcolor[rgb]{ .949,  .949,  .949} 62.11\% & \cellcolor[rgb]{ .949,  .949,  .949} 52.71\% & \cellcolor[rgb]{ .949,  .949,  .949} 60.69\% & \cellcolor[rgb]{ .949,  .949,  .949} 24.20\% & \cellcolor[rgb]{ .949,  .949,  .949} \textbf{49.93\%} \\
          & SparseGPT & 11.2711  & 63.58\% & 53.43\% & 64.33\% & 25.40\% & 51.68\% \\
          & SparseGPT w. OWL & 11.5101  & 63.24\% & 53.07\% & 63.93\% & 23.20\% & 50.86\% \\
          & \cellcolor[rgb]{ .949,  .949,  .949} SparseGPT w. \methodname{} & \cellcolor[rgb]{ .949,  .949,  .949} \textbf{11.2582 } & \cellcolor[rgb]{ .949,  .949,  .949} 64.10\% & \cellcolor[rgb]{ .949,  .949,  .949} 55.43\% & \cellcolor[rgb]{ .949,  .949,  .949} 63.38\% & \cellcolor[rgb]{ .949,  .949,  .949} 25.40\% & \cellcolor[rgb]{ .949,  .949,  .949} \textbf{52.08\%} \\
    \bottomrule
    \end{tabular}}
  \label{tab:appdenix3}%
\end{table}%

\begin{table}[htbp]
  \centering
  \caption{PPL\textdownarrow~ and Zero-shot\textuparrow~ performance of LLaMA-7B at different sparsities.}
  \resizebox*{1\linewidth}{!}{
    \begin{tabular}{clcccccc}
    \toprule
          & LLaMA-7B & PPL\textdownarrow~   & BoolQ & RTE   & WG    & OBQA  & Mean\textuparrow~ \\
    \midrule
    \multirow{6}[2]{*}{\begin{sideways}30\%\end{sideways}} & Magnitude w. OWL & 6.7413  & 71.68\% & 57.04\% & 69.53\% & 31.60\% & 57.46\% \\
          & \cellcolor[rgb]{ .949,  .949,  .949} Magnitude w. \methodname{} & \cellcolor[rgb]{ .949,  .949,  .949} \textbf{6.6769 } & \cellcolor[rgb]{ .949,  .949,  .949} 71.96\% & \cellcolor[rgb]{ .949,  .949,  .949} 61.73\% & \cellcolor[rgb]{ .949,  .949,  .949} 68.98\% & \cellcolor[rgb]{ .949,  .949,  .949} 30.60\% & \cellcolor[rgb]{ .949,  .949,  .949} \textbf{58.32\%} \\
          & Wanda w. OWL & 6.0066  & 76.21\% & 62.45\% & 68.90\% & 32.80\% & 60.09\% \\
          & \cellcolor[rgb]{ .949,  .949,  .949} Wanda w. \methodname{} & \cellcolor[rgb]{ .949,  .949,  .949} \textbf{5.9885 } & \cellcolor[rgb]{ .949,  .949,  .949} 75.81\% & \cellcolor[rgb]{ .949,  .949,  .949} 63.18\% & \cellcolor[rgb]{ .949,  .949,  .949} 69.30\% & \cellcolor[rgb]{ .949,  .949,  .949} 32.80\% & \cellcolor[rgb]{ .949,  .949,  .949} \textbf{60.27\%} \\
          & SparseGPT w. OWL & 5.9512  & 75.57\% & 62.45\% & 69.77\% & 30.80\% & 59.65\% \\
          & \cellcolor[rgb]{ .949,  .949,  .949} SparseGPT w. \methodname{} & \cellcolor[rgb]{ .949,  .949,  .949} \textbf{5.9483 } & \cellcolor[rgb]{ .949,  .949,  .949} 75.72\% & \cellcolor[rgb]{ .949,  .949,  .949} 63.18\% & \cellcolor[rgb]{ .949,  .949,  .949} 70.01\% & \cellcolor[rgb]{ .949,  .949,  .949} 32.60\% & \cellcolor[rgb]{ .949,  .949,  .949} \textbf{60.38\%} \\
    \midrule
    \multirow{6}[2]{*}{\begin{sideways}40\%\end{sideways}} & Magnitude w. OWL & 8.8245  & 67.03\% & 56.68\% & 66.69\% & 31.20\% & 55.40\% \\
          & \cellcolor[rgb]{ .949,  .949,  .949} Magnitude w. \methodname{} & \cellcolor[rgb]{ .949,  .949,  .949} \textbf{8.5035 } & \cellcolor[rgb]{ .949,  .949,  .949} 67.34\% & \cellcolor[rgb]{ .949,  .949,  .949} 59.21\% & \cellcolor[rgb]{ .949,  .949,  .949} 67.32\% & \cellcolor[rgb]{ .949,  .949,  .949} 31.00\% & \cellcolor[rgb]{ .949,  .949,  .949} \textbf{56.22\%} \\
          & Wanda w. OWL & 6.3583  & 74.50\% & 59.21\% & 69.30\% & 31.00\% & 58.50\% \\
          & \cellcolor[rgb]{ .949,  .949,  .949} Wanda w. \methodname{} & \cellcolor[rgb]{ .949,  .949,  .949} \textbf{6.3294 } & \cellcolor[rgb]{ .949,  .949,  .949} 74.37\% & \cellcolor[rgb]{ .949,  .949,  .949} 60.65\% & \cellcolor[rgb]{ .949,  .949,  .949} 68.67\% & \cellcolor[rgb]{ .949,  .949,  .949} 30.40\% & \cellcolor[rgb]{ .949,  .949,  .949} \textbf{58.52\%} \\
          & SparseGPT w. OWL & 6.2543  & 74.74\% & 61.37\% & 68.82\% & 30.20\% & \textbf{58.78\%} \\
          & \cellcolor[rgb]{ .949,  .949,  .949} SparseGPT w. \methodname{} & \cellcolor[rgb]{ .949,  .949,  .949} \textbf{6.2285 } & \cellcolor[rgb]{ .949,  .949,  .949} 74.71\% & \cellcolor[rgb]{ .949,  .949,  .949} 59.93\% & \cellcolor[rgb]{ .949,  .949,  .949} 69.61\% & \cellcolor[rgb]{ .949,  .949,  .949} 29.80\% & \cellcolor[rgb]{ .949,  .949,  .949} 58.51\% \\
    \midrule
    \multirow{6}[2]{*}{\begin{sideways}50\%\end{sideways}} & Magnitude w. OWL & 16.3453  & 58.69\% & 55.23\% & 61.56\% & 27.80\% & 50.82\% \\
          & \cellcolor[rgb]{ .949,  .949,  .949} Magnitude w. \methodname{} & \cellcolor[rgb]{ .949,  .949,  .949} \textbf{15.9755 } & \cellcolor[rgb]{ .949,  .949,  .949} 57.63\% & \cellcolor[rgb]{ .949,  .949,  .949} 57.76\% & \cellcolor[rgb]{ .949,  .949,  .949} 61.40\% & \cellcolor[rgb]{ .949,  .949,  .949} 27.20\% & \cellcolor[rgb]{ .949,  .949,  .949} \textbf{51.00\%} \\
          & Wanda w. OWL & 7.0941  & 73.43\% & 53.43\% & 65.98\% & 30.60\% & 55.86\% \\
          & \cellcolor[rgb]{ .949,  .949,  .949} Wanda w. \methodname{} & \cellcolor[rgb]{ .949,  .949,  .949} \textbf{7.0610 } & \cellcolor[rgb]{ .949,  .949,  .949} 72.91\% & \cellcolor[rgb]{ .949,  .949,  .949} 55.23\% & \cellcolor[rgb]{ .949,  .949,  .949} 67.56\% & \cellcolor[rgb]{ .949,  .949,  .949} 29.80\% & \cellcolor[rgb]{ .949,  .949,  .949} \textbf{56.38\%} \\
          & SparseGPT w. OWL & 6.8785  & 72.84\% & 54.51\% & 67.88\% & 26.00\% & 55.31\% \\
          & \cellcolor[rgb]{ .949,  .949,  .949} SparseGPT w. \methodname{} & \cellcolor[rgb]{ .949,  .949,  .949} \textbf{6.8336 } & \cellcolor[rgb]{ .949,  .949,  .949} 73.30\% & \cellcolor[rgb]{ .949,  .949,  .949} 54.51\% & \cellcolor[rgb]{ .949,  .949,  .949} 68.75\% & \cellcolor[rgb]{ .949,  .949,  .949} 26.80\% & \cellcolor[rgb]{ .949,  .949,  .949} \textbf{55.84\%} \\
    \bottomrule
    \end{tabular}}
  \label{tab:appdenix4}%
\end{table}%

\begin{table}[htbp]
  \centering
  \caption{PPL\textdownarrow~ and Zero-shot\textuparrow~ performance of Vicuna-7B at different sparsities.}
  \resizebox*{1\linewidth}{!}{
    \begin{tabular}{clcccccc}
    \toprule
          & Vicuna-7B & PPL\textdownarrow~   & BoolQ & RTE   & WG    & OBQA  & Mean\textuparrow~ \\
    \midrule
    \multirow{6}[2]{*}{\begin{sideways}30\%\end{sideways}} & Magnitude w. OWL & 8.5803  & 75.41\% & 67.15\% & 66.85\% & 32.00\% & 60.35\% \\
          & \cellcolor[rgb]{ .949,  .949,  .949} Magnitude w. \methodname{} & \cellcolor[rgb]{ .949,  .949,  .949} \textbf{8.1363 } & \cellcolor[rgb]{ .949,  .949,  .949} 75.69\% & \cellcolor[rgb]{ .949,  .949,  .949} 69.68\% & \cellcolor[rgb]{ .949,  .949,  .949} 66.14\% & \cellcolor[rgb]{ .949,  .949,  .949} 32.20\% & \cellcolor[rgb]{ .949,  .949,  .949} \textbf{60.93\%} \\
          & Wanda w. OWL & 7.3325  & 74.40\% & 64.62\% & 68.75\% & 33.00\% & 60.19\% \\
          & \cellcolor[rgb]{ .949,  .949,  .949} Wanda w. \methodname{} & \cellcolor[rgb]{ .949,  .949,  .949} \textbf{7.3159 } & \cellcolor[rgb]{ .949,  .949,  .949} 74.07\% & \cellcolor[rgb]{ .949,  .949,  .949} 65.70\% & \cellcolor[rgb]{ .949,  .949,  .949} 68.51\% & \cellcolor[rgb]{ .949,  .949,  .949} 33.40\% & \cellcolor[rgb]{ .949,  .949,  .949} \textbf{60.42\%} \\
          & SparseGPT w. OWL & 7.2667  & 74.13\% & 63.90\% & 68.67\% & 31.40\% & 59.52\% \\
          & \cellcolor[rgb]{ .949,  .949,  .949} SparseGPT w. \methodname{} & \cellcolor[rgb]{ .949,  .949,  .949} \textbf{7.2622 } & \cellcolor[rgb]{ .949,  .949,  .949} 74.31\% & \cellcolor[rgb]{ .949,  .949,  .949} 64.98\% & \cellcolor[rgb]{ .949,  .949,  .949} 68.03\% & \cellcolor[rgb]{ .949,  .949,  .949} 31.80\% & \cellcolor[rgb]{ .949,  .949,  .949} \textbf{59.78\%} \\
    \midrule
    \multirow{6}[2]{*}{\begin{sideways}40\%\end{sideways}} & Magnitude w. OWL & 11.8580  & 69.60\% & 62.45\% & 64.48\% & 30.20\% & 56.69\% \\
          & \cellcolor[rgb]{ .949,  .949,  .949} Magnitude w. \methodname{} & \cellcolor[rgb]{ .949,  .949,  .949} \textbf{10.8052 } & \cellcolor[rgb]{ .949,  .949,  .949} 73.76\% & \cellcolor[rgb]{ .949,  .949,  .949} 63.18\% & \cellcolor[rgb]{ .949,  .949,  .949} 64.48\% & \cellcolor[rgb]{ .949,  .949,  .949} 29.40\% & \cellcolor[rgb]{ .949,  .949,  .949} \textbf{57.71\%} \\
          & Wanda w. OWL & 7.7982  & 70.52\% & 63.90\% & 66.77\% & 32.00\% & 58.30\% \\
          & \cellcolor[rgb]{ .949,  .949,  .949} Wanda w. \methodname{} & \cellcolor[rgb]{ .949,  .949,  .949} \textbf{7.6862 } & \cellcolor[rgb]{ .949,  .949,  .949} 73.15\% & \cellcolor[rgb]{ .949,  .949,  .949} 67.87\% & \cellcolor[rgb]{ .949,  .949,  .949} 67.25\% & \cellcolor[rgb]{ .949,  .949,  .949} 32.60\% & \cellcolor[rgb]{ .949,  .949,  .949} \textbf{60.22\%} \\
          & SparseGPT w. OWL & 7.5926  & 63.21\% & 68.23\% & 67.56\% & 32.20\% & 57.80\% \\
          & \cellcolor[rgb]{ .949,  .949,  .949} SparseGPT w. \methodname{} & \cellcolor[rgb]{ .949,  .949,  .949} \textbf{7.5080 } & \cellcolor[rgb]{ .949,  .949,  .949} 71.10\% & \cellcolor[rgb]{ .949,  .949,  .949} 63.54\% & \cellcolor[rgb]{ .949,  .949,  .949} 66.69\% & \cellcolor[rgb]{ .949,  .949,  .949} 31.00\% & \cellcolor[rgb]{ .949,  .949,  .949} \textbf{58.08\%} \\
    \midrule
    \multirow{6}[2]{*}{\begin{sideways}50\%\end{sideways}} & Magnitude w. OWL & 21.3578  & 60.67\% & 57.84\% & 60.56\% & 23.00\% & 50.52\% \\
          & \cellcolor[rgb]{ .949,  .949,  .949} Magnitude w. \methodname{} & \cellcolor[rgb]{ .949,  .949,  .949} \textbf{20.7409 } & \cellcolor[rgb]{ .949,  .949,  .949} 60.37\% & \cellcolor[rgb]{ .949,  .949,  .949} 58.84\% & \cellcolor[rgb]{ .949,  .949,  .949} 59.59\% & \cellcolor[rgb]{ .949,  .949,  .949} 23.60\% & \cellcolor[rgb]{ .949,  .949,  .949} \textbf{50.60\%} \\
          & Wanda w. OWL & 8.5731  & 63.15\% & 70.40\% & 66.85\% & 29.60\% & 57.50\% \\
          & \cellcolor[rgb]{ .949,  .949,  .949} Wanda w. \methodname{} & \cellcolor[rgb]{ .949,  .949,  .949} \textbf{8.4428 } & \cellcolor[rgb]{ .949,  .949,  .949} 69.76\% & \cellcolor[rgb]{ .949,  .949,  .949} 69.68\% & \cellcolor[rgb]{ .949,  .949,  .949} 66.30\% & \cellcolor[rgb]{ .949,  .949,  .949} 28.80\% & \cellcolor[rgb]{ .949,  .949,  .949} \textbf{58.63\%} \\
          & SparseGPT w. OWL & 8.2247  & 67.80\% & 67.15\% & 64.72\% & 28.40\% & 57.02\% \\
          & \cellcolor[rgb]{ .949,  .949,  .949} SparseGPT w. \methodname{} & \cellcolor[rgb]{ .949,  .949,  .949} \textbf{8.0762 } & \cellcolor[rgb]{ .949,  .949,  .949} 69.45\% & \cellcolor[rgb]{ .949,  .949,  .949} 68.23\% & \cellcolor[rgb]{ .949,  .949,  .949} 65.67\% & \cellcolor[rgb]{ .949,  .949,  .949} 30.20\% & \cellcolor[rgb]{ .949,  .949,  .949} \textbf{58.39\%} \\
    \bottomrule
    \end{tabular}}
  \label{tab:appdenix5}%
\end{table}%

\begin{table}[htbp]
  \centering
  \caption{PPL\textdownarrow~ and Zero-shot\textuparrow~ performance of LLaMA2-7B at different sparsities.}
  \resizebox*{1\linewidth}{!}{
    \begin{tabular}{clcccccc}
    \toprule
          & LLaMA2-7B & PPL\textdownarrow~   & BoolQ & RTE   & WG    & OBQA  & Mean\textuparrow~ \\
    \midrule
    \multirow{6}[2]{*}{\begin{sideways}30\%\end{sideways}} & Magnitude w. OWL & 6.3403  & 73.33\% & 56.32\% & 69.93\% & 31.20\% & 57.69\% \\
          & \cellcolor[rgb]{ .949,  .949,  .949} Magnitude w. \methodname{} & \cellcolor[rgb]{ .949,  .949,  .949} \textbf{6.3438 } & \cellcolor[rgb]{ .949,  .949,  .949} 72.87\% & \cellcolor[rgb]{ .949,  .949,  .949} 58.12\% & \cellcolor[rgb]{ .949,  .949,  .949} 70.40\% & \cellcolor[rgb]{ .949,  .949,  .949} 32.00\% & \cellcolor[rgb]{ .949,  .949,  .949} \textbf{58.35\%} \\
          & Wanda w. OWL & 5.7712  & 76.79\% & 57.40\% & 68.82\% & 33.20\% & 59.05\% \\
          & \cellcolor[rgb]{ .949,  .949,  .949} Wanda w. \methodname{} & \cellcolor[rgb]{ .949,  .949,  .949} \textbf{5.7554 } & \cellcolor[rgb]{ .949,  .949,  .949} 77.09\% & \cellcolor[rgb]{ .949,  .949,  .949} 57.40\% & \cellcolor[rgb]{ .949,  .949,  .949} 69.38\% & \cellcolor[rgb]{ .949,  .949,  .949} 33.40\% & \cellcolor[rgb]{ .949,  .949,  .949} \textbf{59.32\%} \\
          & SparseGPT w. OWL & 5.7599  & 77.55\% & 57.40\% & 69.85\% & 33.80\% & 59.65\% \\
          & \cellcolor[rgb]{ .949,  .949,  .949} SparseGPT w. \methodname{} & \cellcolor[rgb]{ .949,  .949,  .949} \textbf{5.7535 } & \cellcolor[rgb]{ .949,  .949,  .949} 77.25\% & \cellcolor[rgb]{ .949,  .949,  .949} 61.01\% & \cellcolor[rgb]{ .949,  .949,  .949} 69.53\% & \cellcolor[rgb]{ .949,  .949,  .949} 33.00\% & \cellcolor[rgb]{ .949,  .949,  .949} \textbf{60.20\%} \\
    \midrule
    \multirow{6}[2]{*}{\begin{sideways}40\%\end{sideways}} & Magnitude w. OWL & 8.2723  & 69.94\% & 57.40\% & 68.90\% & 31.20\% & 56.86\% \\
          & \cellcolor[rgb]{ .949,  .949,  .949} Magnitude w. \methodname{} & \cellcolor[rgb]{ .949,  .949,  .949} \textbf{8.1432 } & \cellcolor[rgb]{ .949,  .949,  .949} 69.91\% & \cellcolor[rgb]{ .949,  .949,  .949} 57.76\% & \cellcolor[rgb]{ .949,  .949,  .949} 68.67\% & \cellcolor[rgb]{ .949,  .949,  .949} 32.20\% & \cellcolor[rgb]{ .949,  .949,  .949} \textbf{57.13\%} \\
          & Wanda w. OWL & 6.0827  & 75.81\% & 54.15\% & 68.75\% & 32.20\% & 57.73\% \\
          & \cellcolor[rgb]{ .949,  .949,  .949} Wanda w. \methodname{} & \cellcolor[rgb]{ .949,  .949,  .949} \textbf{6.0702 } & \cellcolor[rgb]{ .949,  .949,  .949} 75.75\% & \cellcolor[rgb]{ .949,  .949,  .949} 59.57\% & \cellcolor[rgb]{ .949,  .949,  .949} 69.06\% & \cellcolor[rgb]{ .949,  .949,  .949} 32.00\% & \cellcolor[rgb]{ .949,  .949,  .949} \textbf{59.09\%} \\
          & SparseGPT w. OWL & 6.0340  & 75.93\% & 57.04\% & 68.51\% & 31.80\% & 58.32\% \\
          & \cellcolor[rgb]{ .949,  .949,  .949} SparseGPT w. \methodname{} & \cellcolor[rgb]{ .949,  .949,  .949} \textbf{6.0306 } & \cellcolor[rgb]{ .949,  .949,  .949} 75.54\% & \cellcolor[rgb]{ .949,  .949,  .949} 56.68\% & \cellcolor[rgb]{ .949,  .949,  .949} 68.43\% & \cellcolor[rgb]{ .949,  .949,  .949} 30.80\% & \cellcolor[rgb]{ .949,  .949,  .949} \textbf{57.86\%} \\
    \midrule
    \multirow{6}[2]{*}{\begin{sideways}50\%\end{sideways}} & Magnitude w. OWL & 15.7421  & 63.88\% & 53.79\% & 62.90\% & 28.60\% & 52.29\% \\
          & \cellcolor[rgb]{ .949,  .949,  .949} Magnitude w. \methodname{} & \cellcolor[rgb]{ .949,  .949,  .949} \textbf{15.3144 } & \cellcolor[rgb]{ .949,  .949,  .949} 64.92\% & \cellcolor[rgb]{ .949,  .949,  .949} 55.96\% & \cellcolor[rgb]{ .949,  .949,  .949} 64.64\% & \cellcolor[rgb]{ .949,  .949,  .949} 28.60\% & \cellcolor[rgb]{ .949,  .949,  .949} \textbf{53.53\%} \\
          & Wanda w. OWL & 6.7954  & 77.22\% & 54.51\% & 68.51\% & 30.40\% & 57.66\% \\
          & \cellcolor[rgb]{ .949,  .949,  .949} Wanda w. \methodname{} & \cellcolor[rgb]{ .949,  .949,  .949} \textbf{6.7959 } & \cellcolor[rgb]{ .949,  .949,  .949} 76.18\% & \cellcolor[rgb]{ .949,  .949,  .949} 54.15\% & \cellcolor[rgb]{ .949,  .949,  .949} 69.40\% & \cellcolor[rgb]{ .949,  .949,  .949} 31.20\% & \cellcolor[rgb]{ .949,  .949,  .949} \textbf{57.73\%} \\
          & SparseGPT w. OWL & 6.6388  & 71.50\% & 56.32\% & 68.19\% & 28.40\% & 56.10\% \\
          & \cellcolor[rgb]{ .949,  .949,  .949} SparseGPT w. \methodname{} & \cellcolor[rgb]{ .949,  .949,  .949} \textbf{6.5843 } & \cellcolor[rgb]{ .949,  .949,  .949} 68.84\% & \cellcolor[rgb]{ .949,  .949,  .949} 63.54\% & \cellcolor[rgb]{ .949,  .949,  .949} 67.96\% & \cellcolor[rgb]{ .949,  .949,  .949} 28.60\% & \cellcolor[rgb]{ .949,  .949,  .949} \textbf{57.23\%} \\
    \bottomrule
    \end{tabular}}
  \label{tab:appdenix6}%
\end{table}%

\begin{table}[htbp]
  \centering
  \caption{PPL\textdownarrow~ and Zero-shot\textuparrow~ performance of LLaMA3.2-3B at different sparsities.}
  \resizebox*{1\linewidth}{!}{
    \begin{tabular}{clcccccc}
    \toprule
          & LLaMA3.2-3B & PPL\textdownarrow~   & BoolQ & RTE   & WG    & OBQA  & Mean\textuparrow~ \\
    \midrule
    \multirow{6}[2]{*}{\begin{sideways}30\%\end{sideways}} & Magnitude w. OWL & 9.9054  & 64.71\% & 57.40\% & 66.93\% & 30.20\% & 54.81\% \\
          & \cellcolor[rgb]{ .949,  .949,  .949} Magnitude w. \methodname{} & \cellcolor[rgb]{ .949,  .949,  .949} \textbf{9.9530 } & \cellcolor[rgb]{ .949,  .949,  .949} 68.20\% & \cellcolor[rgb]{ .949,  .949,  .949} 57.40\% & \cellcolor[rgb]{ .949,  .949,  .949} 68.11\% & \cellcolor[rgb]{ .949,  .949,  .949} 30.00\% & \cellcolor[rgb]{ .949,  .949,  .949} \textbf{55.93\%} \\
          & Wanda w. OWL & 8.4789  & 71.87\% & 50.54\% & 70.01\% & 31.60\% & 56.00\% \\
          & \cellcolor[rgb]{ .949,  .949,  .949} Wanda w. \methodname{} & \cellcolor[rgb]{ .949,  .949,  .949} \textbf{8.4111 } & \cellcolor[rgb]{ .949,  .949,  .949} 73.30\% & \cellcolor[rgb]{ .949,  .949,  .949} 52.35\% & \cellcolor[rgb]{ .949,  .949,  .949} 70.40\% & \cellcolor[rgb]{ .949,  .949,  .949} 31.40\% & \cellcolor[rgb]{ .949,  .949,  .949} \textbf{56.86\%} \\
          & SparseGPT w. OWL & 8.3514  & 73.76\% & 54.15\% & 70.09\% & 31.80\% & \textbf{57.45\%} \\
          & \cellcolor[rgb]{ .949,  .949,  .949} SparseGPT w. \methodname{} & \cellcolor[rgb]{ .949,  .949,  .949} \textbf{8.3202 } & \cellcolor[rgb]{ .949,  .949,  .949} 73.67\% & \cellcolor[rgb]{ .949,  .949,  .949} 54.51\% & \cellcolor[rgb]{ .949,  .949,  .949} 70.24\% & \cellcolor[rgb]{ .949,  .949,  .949} 31.00\% & \cellcolor[rgb]{ .949,  .949,  .949} 57.36\% \\
    \midrule
    \multirow{6}[2]{*}{\begin{sideways}40\%\end{sideways}} & Magnitude w. OWL & \textbf{15.8035 } & 54.28\% & 54.87\% & 63.38\% & 25.40\% & \textbf{49.48\%} \\
          & \cellcolor[rgb]{ .949,  .949,  .949} Magnitude w. \methodname{} & \cellcolor[rgb]{ .949,  .949,  .949} 16.2707  & \cellcolor[rgb]{ .949,  .949,  .949} 53.85\% & \cellcolor[rgb]{ .949,  .949,  .949} 53.79\% & \cellcolor[rgb]{ .949,  .949,  .949} 64.25\% & \cellcolor[rgb]{ .949,  .949,  .949} 24.80\% & \cellcolor[rgb]{ .949,  .949,  .949} 49.17\% \\
          & Wanda w. OWL & 9.6285  & 68.44\% & 55.96\% & 67.72\% & 29.00\% & 55.28\% \\
          & \cellcolor[rgb]{ .949,  .949,  .949} Wanda w. \methodname{} & \cellcolor[rgb]{ .949,  .949,  .949} \textbf{9.4400 } & \cellcolor[rgb]{ .949,  .949,  .949} 70.24\% & \cellcolor[rgb]{ .949,  .949,  .949} 55.23\% & \cellcolor[rgb]{ .949,  .949,  .949} 68.11\% & \cellcolor[rgb]{ .949,  .949,  .949} 30.00\% & \cellcolor[rgb]{ .949,  .949,  .949} \textbf{55.90\%} \\
          & SparseGPT w. OWL & 9.1586  & 74.01\% & 49.46\% & 69.53\% & 29.20\% & 55.55\% \\
          & \cellcolor[rgb]{ .949,  .949,  .949} SparseGPT w. \methodname{} & \cellcolor[rgb]{ .949,  .949,  .949} \textbf{9.0932 } & \cellcolor[rgb]{ .949,  .949,  .949} 71.99\% & \cellcolor[rgb]{ .949,  .949,  .949} 51.99\% & \cellcolor[rgb]{ .949,  .949,  .949} 68.43\% & \cellcolor[rgb]{ .949,  .949,  .949} 30.00\% & \cellcolor[rgb]{ .949,  .949,  .949} \textbf{55.60\%} \\
    \midrule
    \multirow{6}[2]{*}{\begin{sideways}50\%\end{sideways}} & Magnitude w. OWL & 99.2163  & 41.28\% & 51.62\% & 54.78\% & 16.80\% & 41.12\% \\
          & \cellcolor[rgb]{ .949,  .949,  .949} Magnitude w. \methodname{} & \cellcolor[rgb]{ .949,  .949,  .949} \textbf{89.4710 } & \cellcolor[rgb]{ .949,  .949,  .949} 44.07\% & \cellcolor[rgb]{ .949,  .949,  .949} 51.99\% & \cellcolor[rgb]{ .949,  .949,  .949} 56.99\% & \cellcolor[rgb]{ .949,  .949,  .949} 17.60\% & \cellcolor[rgb]{ .949,  .949,  .949} \textbf{42.66\%} \\
          & Wanda w. OWL & 12.6735  & 63.98\% & 53.07\% & 64.56\% & 24.80\% & 51.60\% \\
          & \cellcolor[rgb]{ .949,  .949,  .949} Wanda w. \methodname{} & \cellcolor[rgb]{ .949,  .949,  .949} \textbf{12.1578 } & \cellcolor[rgb]{ .949,  .949,  .949} 69.88\% & \cellcolor[rgb]{ .949,  .949,  .949} 55.96\% & \cellcolor[rgb]{ .949,  .949,  .949} 65.27\% & \cellcolor[rgb]{ .949,  .949,  .949} 24.40\% & \cellcolor[rgb]{ .949,  .949,  .949} \textbf{53.88\%} \\
          & SparseGPT w. OWL & 10.9828  & 70.28\% & 49.46\% & 66.30\% & 24.80\% & 52.71\% \\
          & \cellcolor[rgb]{ .949,  .949,  .949} SparseGPT w. \methodname{} & \cellcolor[rgb]{ .949,  .949,  .949} \textbf{10.8495 } & \cellcolor[rgb]{ .949,  .949,  .949} 70.46\% & \cellcolor[rgb]{ .949,  .949,  .949} 51.62\% & \cellcolor[rgb]{ .949,  .949,  .949} 65.98\% & \cellcolor[rgb]{ .949,  .949,  .949} 25.20\% & \cellcolor[rgb]{ .949,  .949,  .949} \textbf{53.32\%} \\
    \bottomrule
    \end{tabular}}
  \label{tab:appdenix7}%
\end{table}%

\begin{table}[htbp]
  \centering
  \caption{PPL\textdownarrow~ and Zero-shot\textuparrow~ performance of OPT-2.7B at different sparsities.}
  \resizebox*{1\linewidth}{!}{
    \begin{tabular}{clcccccc}
    \toprule
          & OPT-2.7B & PPL\textdownarrow~   & BoolQ & RTE   & WG    & OBQA  & Mean\textuparrow~ \\
    \midrule
    \multirow{6}[2]{*}{\begin{sideways}30\%\end{sideways}} & Magnitude w. OWL & 15.0581  & 46.02\% & 54.51\% & 57.70\% & 24.00\% & 45.56\% \\
          & \cellcolor[rgb]{ .949,  .949,  .949} Magnitude w. \methodname{} & \cellcolor[rgb]{ .949,  .949,  .949} \textbf{14.4552 } & \cellcolor[rgb]{ .949,  .949,  .949} 48.87\% & \cellcolor[rgb]{ .949,  .949,  .949} 52.71\% & \cellcolor[rgb]{ .949,  .949,  .949} 59.67\% & \cellcolor[rgb]{ .949,  .949,  .949} 22.60\% & \cellcolor[rgb]{ .949,  .949,  .949} \textbf{45.96\%} \\
          & Wanda w. OWL & 12.3628  & 65.50\% & 51.26\% & 60.46\% & 24.20\% & 50.36\% \\
          & \cellcolor[rgb]{ .949,  .949,  .949} Wanda w. \methodname{} & \cellcolor[rgb]{ .949,  .949,  .949} \textbf{12.2264 } & \cellcolor[rgb]{ .949,  .949,  .949} 66.27\% & \cellcolor[rgb]{ .949,  .949,  .949} 51.26\% & \cellcolor[rgb]{ .949,  .949,  .949} 60.06\% & \cellcolor[rgb]{ .949,  .949,  .949} 24.20\% & \cellcolor[rgb]{ .949,  .949,  .949} \textbf{50.45\%} \\
          & SparseGPT w. OWL & 12.2349  & 64.28\% & 52.35\% & 60.30\% & 24.60\% & 50.38\% \\
          & \cellcolor[rgb]{ .949,  .949,  .949} SparseGPT w. \methodname{} & \cellcolor[rgb]{ .949,  .949,  .949} \textbf{12.1657 } & \cellcolor[rgb]{ .949,  .949,  .949} 66.54\% & \cellcolor[rgb]{ .949,  .949,  .949} 53.79\% & \cellcolor[rgb]{ .949,  .949,  .949} 60.14\% & \cellcolor[rgb]{ .949,  .949,  .949} 22.80\% & \cellcolor[rgb]{ .949,  .949,  .949} \textbf{50.82\%} \\
    \midrule
    \multirow{6}[2]{*}{\begin{sideways}40\%\end{sideways}} & Magnitude w. OWL & 22.9237  & 44.65\% & 52.35\% & 58.17\% & 22.40\% & 44.39\% \\
          & \cellcolor[rgb]{ .949,  .949,  .949} Magnitude w. \methodname{} & \cellcolor[rgb]{ .949,  .949,  .949} \textbf{22.8636 } & \cellcolor[rgb]{ .949,  .949,  .949} 44.01\% & \cellcolor[rgb]{ .949,  .949,  .949} 51.99\% & \cellcolor[rgb]{ .949,  .949,  .949} 58.33\% & \cellcolor[rgb]{ .949,  .949,  .949} 23.80\% & \cellcolor[rgb]{ .949,  .949,  .949} \textbf{44.53\%} \\
          & Wanda w. OWL & 13.0116  & 62.48\% & 52.71\% & 58.72\% & 23.60\% & \textbf{49.38\%} \\
          & \cellcolor[rgb]{ .949,  .949,  .949} Wanda w. \methodname{} & \cellcolor[rgb]{ .949,  .949,  .949} \textbf{12.7347 } & \cellcolor[rgb]{ .949,  .949,  .949} 62.97\% & \cellcolor[rgb]{ .949,  .949,  .949} 51.62\% & \cellcolor[rgb]{ .949,  .949,  .949} 58.80\% & \cellcolor[rgb]{ .949,  .949,  .949} 24.00\% & \cellcolor[rgb]{ .949,  .949,  .949} 49.35\% \\
          & SparseGPT w. OWL & 12.5431  & 64.62\% & 53.07\% & 59.19\% & 23.40\% & 50.07\% \\
          & \cellcolor[rgb]{ .949,  .949,  .949} SparseGPT w. \methodname{} & \cellcolor[rgb]{ .949,  .949,  .949} \textbf{12.3850 } & \cellcolor[rgb]{ .949,  .949,  .949} 65.93\% & \cellcolor[rgb]{ .949,  .949,  .949} 53.07\% & \cellcolor[rgb]{ .949,  .949,  .949} 59.67\% & \cellcolor[rgb]{ .949,  .949,  .949} 24.00\% & \cellcolor[rgb]{ .949,  .949,  .949} \textbf{50.67\%} \\
    \midrule
    \multirow{6}[2]{*}{\begin{sideways}50\%\end{sideways}} & Magnitude w. OWL & 207.0598  & 38.17\% & 52.71\% & 53.67\% & 21.20\% & 41.44\% \\
          & \cellcolor[rgb]{ .949,  .949,  .949} Magnitude w. \methodname{} & \cellcolor[rgb]{ .949,  .949,  .949} \textbf{158.2674 } & \cellcolor[rgb]{ .949,  .949,  .949} 38.59\% & \cellcolor[rgb]{ .949,  .949,  .949} 52.35\% & \cellcolor[rgb]{ .949,  .949,  .949} 53.51\% & \cellcolor[rgb]{ .949,  .949,  .949} 22.40\% & \cellcolor[rgb]{ .949,  .949,  .949} \textbf{41.71\%} \\
          & Wanda w. OWL & 14.7622  & 61.90\% & 52.35\% & 57.70\% & 20.80\% & 48.18\% \\
          & \cellcolor[rgb]{ .949,  .949,  .949} Wanda w. \methodname{} & \cellcolor[rgb]{ .949,  .949,  .949} \textbf{14.0042 } & \cellcolor[rgb]{ .949,  .949,  .949} 62.29\% & \cellcolor[rgb]{ .949,  .949,  .949} 51.62\% & \cellcolor[rgb]{ .949,  .949,  .949} 58.41\% & \cellcolor[rgb]{ .949,  .949,  .949} 21.60\% & \cellcolor[rgb]{ .949,  .949,  .949} \textbf{48.48\%} \\
          & SparseGPT w. OWL & 13.3490  & 62.87\% & 51.62\% & 58.80\% & 23.20\% & 49.12\% \\
          & \cellcolor[rgb]{ .949,  .949,  .949} SparseGPT w. \methodname{} & \cellcolor[rgb]{ .949,  .949,  .949} \textbf{12.9768 } & \cellcolor[rgb]{ .949,  .949,  .949} 63.30\% & \cellcolor[rgb]{ .949,  .949,  .949} 52.71\% & \cellcolor[rgb]{ .949,  .949,  .949} 59.43\% & \cellcolor[rgb]{ .949,  .949,  .949} 22.60\% & \cellcolor[rgb]{ .949,  .949,  .949} \textbf{49.51\%} \\
    \bottomrule
    \end{tabular}}
  \label{tab:appdenix8}%
\end{table}%

\begin{table}[htbp]
  \centering
  \caption{PPL\textdownarrow~ and Zero-shot\textuparrow~ performance of OPT-6.7B at different sparsities.}
  \resizebox*{1\linewidth}{!}{
    \begin{tabular}{clcccccc}
    \toprule
          & OPT-6.7B & PPL\textdownarrow~   & BoolQ & RTE   & WG    & OBQA  & Mean\textuparrow~ \\
    \midrule
    \multirow{6}[2]{*}{\begin{sideways}30\%\end{sideways}} & Magnitude w. OWL & 12.8504  & 49.79\% & 53.43\% & 59.98\% & 26.00\% & 47.30\% \\
          & \cellcolor[rgb]{ .949,  .949,  .949} Magnitude w. \methodname{} & \cellcolor[rgb]{ .949,  .949,  .949} \textbf{12.3753 } & \cellcolor[rgb]{ .949,  .949,  .949} 52.78\% & \cellcolor[rgb]{ .949,  .949,  .949} 53.79\% & \cellcolor[rgb]{ .949,  .949,  .949} 60.06\% & \cellcolor[rgb]{ .949,  .949,  .949} 26.20\% & \cellcolor[rgb]{ .949,  .949,  .949} \textbf{48.21\%} \\
          & Wanda w. OWL & 10.7365  & 65.72\% & 54.15\% & 63.93\% & 27.60\% & 52.85\% \\
          & \cellcolor[rgb]{ .949,  .949,  .949} Wanda w. \methodname{} & \cellcolor[rgb]{ .949,  .949,  .949} \textbf{10.6790 } & \cellcolor[rgb]{ .949,  .949,  .949} 66.48\% & \cellcolor[rgb]{ .949,  .949,  .949} 53.43\% & \cellcolor[rgb]{ .949,  .949,  .949} 64.01\% & \cellcolor[rgb]{ .949,  .949,  .949} 28.60\% & \cellcolor[rgb]{ .949,  .949,  .949} \textbf{53.13\%} \\
          & SparseGPT w. OWL & 10.9449  & 67.77\% & 53.79\% & 63.77\% & 27.00\% & 53.08\% \\
          & \cellcolor[rgb]{ .949,  .949,  .949} SparseGPT w. \methodname{} & \cellcolor[rgb]{ .949,  .949,  .949} \textbf{10.8853 } & \cellcolor[rgb]{ .949,  .949,  .949} 67.89\% & \cellcolor[rgb]{ .949,  .949,  .949} 54.15\% & \cellcolor[rgb]{ .949,  .949,  .949} 63.77\% & \cellcolor[rgb]{ .949,  .949,  .949} 27.00\% & \cellcolor[rgb]{ .949,  .949,  .949} \textbf{53.20\%} \\
    \midrule
    \multirow{6}[2]{*}{\begin{sideways}40\%\end{sideways}} & Magnitude w. OWL & 20.4357  & 43.00\% & 52.71\% & 58.48\% & 25.40\% & 44.90\% \\
          & \cellcolor[rgb]{ .949,  .949,  .949} Magnitude w. \methodname{} & \cellcolor[rgb]{ .949,  .949,  .949} \textbf{18.7027 } & \cellcolor[rgb]{ .949,  .949,  .949} 44.34\% & \cellcolor[rgb]{ .949,  .949,  .949} 52.71\% & \cellcolor[rgb]{ .949,  .949,  .949} 58.72\% & \cellcolor[rgb]{ .949,  .949,  .949} 25.60\% & \cellcolor[rgb]{ .949,  .949,  .949} \textbf{45.34\%} \\
          & Wanda w. OWL & 11.2408  & 62.35\% & 52.35\% & 62.90\% & 26.80\% & \textbf{51.10\%} \\
          & \cellcolor[rgb]{ .949,  .949,  .949} Wanda w. \methodname{} & \cellcolor[rgb]{ .949,  .949,  .949} \textbf{11.1051 } & \cellcolor[rgb]{ .949,  .949,  .949} 62.94\% & \cellcolor[rgb]{ .949,  .949,  .949} 52.35\% & \cellcolor[rgb]{ .949,  .949,  .949} 62.67\% & \cellcolor[rgb]{ .949,  .949,  .949} 26.20\% & \cellcolor[rgb]{ .949,  .949,  .949} 51.04\% \\
          & SparseGPT w. OWL & 11.0805  & 64.95\% & 53.79\% & 64.25\% & 26.60\% & 52.40\% \\
          & \cellcolor[rgb]{ .949,  .949,  .949} SparseGPT w. \methodname{} & \cellcolor[rgb]{ .949,  .949,  .949} \textbf{10.9279 } & \cellcolor[rgb]{ .949,  .949,  .949} 66.36\% & \cellcolor[rgb]{ .949,  .949,  .949} 54.51\% & \cellcolor[rgb]{ .949,  .949,  .949} 63.85\% & \cellcolor[rgb]{ .949,  .949,  .949} 26.00\% & \cellcolor[rgb]{ .949,  .949,  .949} \textbf{52.68\%} \\
    \midrule
    \multirow{6}[2]{*}{\begin{sideways}50\%\end{sideways}} & Magnitude w. OWL & 363.6955  & 38.13\% & 52.71\% & 55.64\% & 21.40\% & 41.97\% \\
          & \cellcolor[rgb]{ .949,  .949,  .949} Magnitude w. \methodname{} & \cellcolor[rgb]{ .949,  .949,  .949} \textbf{308.4234 } & \cellcolor[rgb]{ .949,  .949,  .949} 39.08\% & \cellcolor[rgb]{ .949,  .949,  .949} 52.71\% & \cellcolor[rgb]{ .949,  .949,  .949} 55.67\% & \cellcolor[rgb]{ .949,  .949,  .949} 21.20\% & \cellcolor[rgb]{ .949,  .949,  .949} \textbf{42.17\%} \\
          & Wanda w. OWL & 12.3921  & 62.20\% & 52.71\% & 58.64\% & 23.80\% & 49.34\% \\
          & \cellcolor[rgb]{ .949,  .949,  .949} Wanda w. \methodname{} & \cellcolor[rgb]{ .949,  .949,  .949} \textbf{12.0065 } & \cellcolor[rgb]{ .949,  .949,  .949} 62.11\% & \cellcolor[rgb]{ .949,  .949,  .949} 52.71\% & \cellcolor[rgb]{ .949,  .949,  .949} 60.69\% & \cellcolor[rgb]{ .949,  .949,  .949} 24.20\% & \cellcolor[rgb]{ .949,  .949,  .949} \textbf{49.93\%} \\
          & SparseGPT w. OWL & 11.5101  & 63.24\% & 53.07\% & 63.93\% & 23.20\% & 50.86\% \\
          & \cellcolor[rgb]{ .949,  .949,  .949} SparseGPT w. \methodname{} & \cellcolor[rgb]{ .949,  .949,  .949} \textbf{11.2582 } & \cellcolor[rgb]{ .949,  .949,  .949} 64.10\% & \cellcolor[rgb]{ .949,  .949,  .949} 55.43\% & \cellcolor[rgb]{ .949,  .949,  .949} 63.38\% & \cellcolor[rgb]{ .949,  .949,  .949} 25.40\% & \cellcolor[rgb]{ .949,  .949,  .949} \textbf{52.08\%} \\
    \bottomrule
    \end{tabular}}
  \label{tab:appdenix9}%
\end{table}%

\end{document}